\documentclass[10pt,twocolumn,letterpaper]{article}

\usepackage{cvpr}              
\usepackage{algorithmic}
\usepackage{algorithm}

\usepackage[dvipsnames]{xcolor}
\usepackage{bbding}
\usepackage{enumerate}
\usepackage{enumitem}
\usepackage{svg}

\definecolor{cvprblue}{rgb}{0.21,0.49,0.74}
\usepackage[pagebackref,breaklinks,colorlinks,citecolor=cvprblue]{hyperref}

\def\paperID{62} 
\def\confName{CVPR}
\def\confYear{2025}
\usepackage{inputenc}

\usepackage{amsthm}
\usepackage{booktabs}
\usepackage{algorithm}
\usepackage{algorithmic}
\usepackage[switch]{lineno}
\usepackage{adjustbox}
\usepackage{multirow} 
\usepackage{makecell}
\usepackage{color}
\usepackage{caption}
\usepackage{subcaption}
\usepackage{makecell}
\usepackage[accsupp]{axessibility}
\usepackage{amssymb}

\title{FusionNet: Multi-model Linear Fusion Framework\\for Low-light Image Enhancement}
\author{Kangbiao Shi$^1$\thanks{These authors contribute equally.}
~~~~~Yixu Feng$^{1*}$~~~~~Tao Hu$^1$\\Yu Cao$^3$~~~~~Peng Wu$^1$~~~~~Yijin Liang$^4$~~~~~Yanning Zhang$^1$~~~~~
Qingsen Yan$^{1,2}$\thanks{Qingsen Yan (qingsenyan@nwpu.edu.cn) is the corresponding author. This work is supported by NSFC of China under Grant 62301432 and Grant 6230624, the Natural
Science Basic Research Program of Shaanxi under Grant 2023-JC-QN-0685
and Grant QCYRCXM-2023-057, the Fundamental Research
Funds for Central Universities, and Guangdong Basic and Applied Basic Research Foundation 2025A1515011119.}
\\
$^1$Northwestern Polytechnical University\\
$^2$Shenzhen Research Institute of Northwestern Polytechnical University\\
$^3$Xi’an Institute of Optics and Precision Mechanics of CAS\\
$^4$Shanghai Institute of Satellite Engineering, Shanghai 201109, China
}
\begin{document}

\maketitle

\begin{abstract}
The advent of Deep Neural Networks (DNNs) has driven remarkable progress in low-light image enhancement (LLIE), with diverse architectures (\eg, CNNs and Transformers) and color spaces (\eg, sRGB, HSV, HVI) yielding impressive results. Recent efforts have sought to leverage the complementary strengths of these paradigms, offering promising solutions to enhance performance across varying degradation scenarios. However, existing fusion strategies are hindered by challenges such as parameter explosion, optimization instability, and feature misalignment, limiting further improvements. To overcome these issues, we introduce FusionNet, a novel multi-model linear fusion framework that operates in parallel to effectively capture global and local features across diverse color spaces. By incorporating a linear fusion strategy underpinned by Hilbert space theoretical guarantees, FusionNet mitigates network collapse and reduces excessive training costs. Our method achieved 1st place in the CVPR2025 NTIRE Low Light Enhancement Challenge. Extensive experiments conducted on synthetic and real-world benchmark datasets demonstrate that the proposed method significantly outperforms state-of-the-art methods in terms of both quantitative and qualitative results, delivering robust enhancement under diverse low-light conditions.
\end{abstract}
\begin{figure}
\includegraphics[width=\linewidth,scale=1]{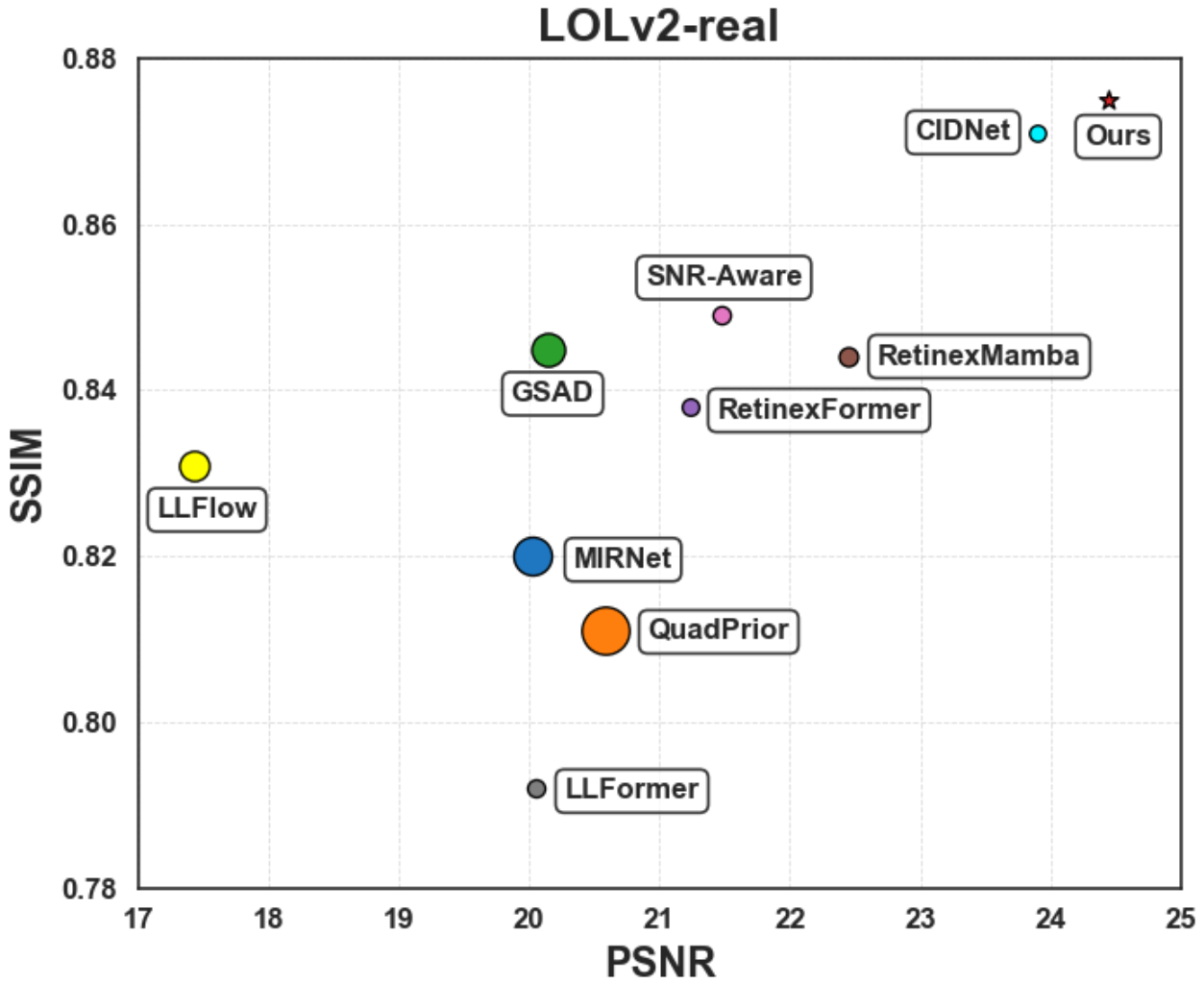}
\caption{Comparison with recent SOTA methods in LOLv2-real \cite{lol_v2} dataset. We choose PSNR$\uparrow$ and SSIM$\uparrow$ for the measure metrics. The size of the circles indicates the FLOPs of each model. The result shows that proposed FusionNet achived the best performance among these methods.}
\label{fig:1}
\end{figure}   
\renewcommand{\thefootnote}{}
\section{Introduction}
\label{sec:intro}
Images captured in low-light conditions often suffer from significant image quality degradation, characterized by poor visibility, low contrast, and high noise levels. 
These visual flaws not only impede human perceptual analysis but also pose significant barriers to vision-based applications, such as object detection and semantic segmentation. 
Therefore, developing advanced Low-Light Image Enhancement (LLIE) methods is crucial to improve the subjective quality of images and ensure the operational reliability of vision systems under adverse lighting conditions.

LLIE aims to restore illumination in images while simultaneously correcting degradations such as noise, artifacts, and color distortion. 
%
%
Traditional methods \cite{ibrahim2007brightness,abdullah2007dynamic,guo2016lime} typically rely on hand-crafted optimization based on predefined priors and assumptions. However, due to the challenge of adapting these priors to diverse environments, these methods struggle to achieve robust performance across varying lighting conditions.
In recent years, the rise of Deep Neural Networks (DNNs) has brought significant advancements due to the powerful nonlinear representation ability. Numerous DNN-based LLIE methods have emerged, which have two different classifications: (1) Convolutional Neural Networks (CNNs) \cite{wei1808deep,zhang2019kindling} or Vision Transformers (ViTs) \cite{snr_net,cai2023retinexformer}, and (2) sRGB-based \cite{guo2020zero,zhou2022lednet} or other color-model-based method \cite{yan2025hvi,guo2023low,zhou2025low}.
In the first network classification, CNN-based methods have achieved remarkable progress due to the powerful nonlinear representation ability of neural networks. Also, transformer-based methods have been introduced as an alternative to CNNs, which excel at capturing global contextual information by leveraging self-attention mechanisms. 
In the second color space classification, sRGB enables direct brightness adjustment but suffers from coupled color-brightness interference. While other color-model-based method try to decouple brightness and color from sRGB color space, yet exhibits higher data-driven sensitivity (Retinex \cite{wu2022uretinex}, HVI \cite{yan2025hvi}) and discontinuity of color representation (HSV \cite{zhou2025low}).

Methods referred to in these two classifications each have their strengths and weaknesses. As a result, some methods have already attempted to integrate different networks and color spaces to improve overall performance \cite{feng2024difflight,snr_net,9426457,du2024multicolor}.
These methods have demonstrated that integrating the advantages of different networks and color spaces can enhance model robustness across diverse scenarios, offering a promising solution for low-light image enhancement under varying degrees of degradation.
However, existing fusion methods face critical limitations: serially connect architectures suffer from parameter explosion and optimization instability due to sequential single-stage training \cite{Li_2023_CVPR}; multi-stage serial pipelines incur escalating iteration costs and convergence uncertainties by freezing prior modules incrementally \cite{Yang_2024_CVPR}; hybrid parallel-serial frameworks inherit multi-stage inefficiencies while introducing feature misalignment from concatenated outputs \cite{Li_2023_CVPR}. So, is there a simpler fusion strategy that can effectively avoid these issues?

Based on the above analysis, we propose FusionNet, a multi-model linear fusion framework that enhances different models which adopts a fully parallel execution structure where networks operate independently without mutual interference.
Specifically, FusionNet consists of three different method: (1) CIDNet \cite{yan2025hvi} as the representative of an HVI-based transformer method, (2) an sRGB-based method Retinexformer \cite{cai2023retinexformer}, and (3) CNN/RGB-based ESDNet \cite{yu2022towards}, to extract both global and local information from the sRGB and HVI space. 
During training, all three models undergo independent optimization. For inference, our proposed linear fusion strategy executes a parallel integration, by combining these methods in a linear function to generate a specific output image.
More importantly, by leveraging Hilbert space, we theoretically demonstrate that this method can effectively mitigate network collapse and excessive training time issues present in other fusion strategies.
The proposed FusionNet achieved \textbf{rank \#1} in the CVPR 2025 NTIRE Workshop Low-Light Image Enhancement Challenge. Experimental results across various challenging scenarios demonstrate that our method attains SOTA performance in both quantitative and qualitative evaluations for the LLIE task. The main contributions are summarized as follows:
\begin{itemize}
    \item We introduce a novel linear fusion framework that integrates multiple models to address the limitations of single-method approaches in low-light image enhancement.
    \item We establish linear fusion with hilbert space theoretical guarantees, resolving network collapse and training inefficiencies plaguing prior fusion paradigms.
    \item Extensive experimental evaluations validate that our approach significantly outperforms SOTA methods across multiple metrics on challenging low-light datasets.
\end{itemize}


\section{Related Work}
\label{sec:relative}
\subsection{Deep Learning Enhancement}
\textbf{CNN-based Methods.} 
CNN-based LLIE methods can usually remove severe local noise caused by low-light conditions while preserving the original structure and fine details of the image as much as possible.
Lore \textit{et al.}\cite{Lore2017LLNet} introduced a stacked sparse denoising auto encoder to simultaneously brighten and denoise images, which is the first CNN-based method of LLIE task.
Zhou \textit{et al.} \cite{zhou2022lednet} focuses on joint low-light enhancement and deblurring, introducing a large-scale dataset LOL-Blur and demonstrating effectiveness on both synthetic and real-world datasets.
However, these models struggle with global contrast enhancement, often leading to overexposure and suboptimal enhancement results.

\noindent\textbf{Transformers.} 
More recently, Transformer-based methods have been introduced as an alternative to CNNs, which excel at capturing global contextual information.
SNR-Aware \cite{snr_net} presents a collective Signal-to-Noise-Ratio-aware transformer network to dynamically enhance pixels with spatial-varying operations, which could reduce the color bias and noise in sRGB domain.
LLformer \cite{wang2023ultra} model achieves high-resolution image restoration through an effective Transformer architecture.
Retinexformer \cite{cai2023retinexformer} presents a single-stage low-light image enhancement framework based on Retinex theory, which uses illumination-guided Transformers to significantly boost image enhancement results.
These models exhibit strong global enhancement capabilities but tend to introduce local artifacts, which degrade visual quality.

\subsection{Color Space} 
\textbf{sRGB.} As a standard space for digital imaging, sRGB has the advantage of device-independent color reproduction. 
Currently, most sRGB-based methods employ a complex network to directly map the input to the output domain \cite{zhou2023pyramid,wang2023fourllie,wang2022low,jiang2021enlightengan}.
Nevertheless, image brightness and color exhibit a strong interdependence with the three channels in sRGB \cite{gevers2012color}.
Therefore, some models have attempted to leverage Retinex theory to decouple RGB images into luminance and color components.
KinD \cite{zhang2019kindling} and KinD++ \cite{zhang2021beyond} adopt the decomposition and adjustment paradigm and use the CNN to learn the mapping in both decomposition and adjustment.
Diff-Retinex \cite{yi2023diff} integrate the superiority of attention in Transformer and meticulously design a Retinex Transformer decomposition network (TDN) to decompose the image into illumination and reflectance maps.
This Retinex-based decomposition from sRGB typically requires a network with a large number of parameters or high computational cost \cite{liu2021retinex}, significantly impacting performance. 
\textbf{HSV and HVI.} 
The HSV color space breaks down color information into intuitive hue, saturation, and value components, which decouples brightness (value-axis) and color (HS plane) of the image from sRGB channels.
Zhou \textit{et al.} \cite{zhou2025low} use the HSV space to decompose the image for initial processing, and then applying Bayesian rules to fuse saturation and value, converts the three components back to the RGB space for a rough enhanced image. 
However this HSV-based low-light image enhancement inevitably introduces red-black artifacts \cite{yan2025hvi}.
Therefore, Yan \textit{et al.} introduced the HVI color space by polarization HS plane and conducted a $\mathbf{C}_k$ formula to resolve the red and black discontinuity issue.
While Unlike sRGB, which is an absolute three-axis independent space, HVI is a data-drvien color space with a trainable parameter $k$ that could lack such strict independence. 
As a result, when switching datasets or scenes, HVI's generalization ability may be compromised.

\subsection{Fusion Strategy}
\textbf{Multi-network fusion strategy.} It is a common multi-network fusion strategy frequently used by teams in NTIRE competitions \cite{NTIRE,Li_2023_CVPR}. However, it typically does not achieve high performance metrics. This is because cascading multiple network structures deepens the overall architecture, which increases the risk of network collapse, where small parameter variations can cause significant performance fluctuations.\\ \textbf{Multi-stage serial and parallel-stage network.} These types of fusion strategy widely used in low-light enhancement models (\eg FourLLIE \cite{wang2023fourllie}, DiffLight \cite{feng2024difflight}, and GSAD \cite{hou2024global}), where the model is divided into multiple stages, with each new stage freezing the parameters of the previous one. Experimental results have proven the effectiveness of this approach. However, the training process requires a large number of iterations, leading to significant time costs.

\begin{figure*}[h]
    \centering
    \includegraphics[width=1\linewidth]{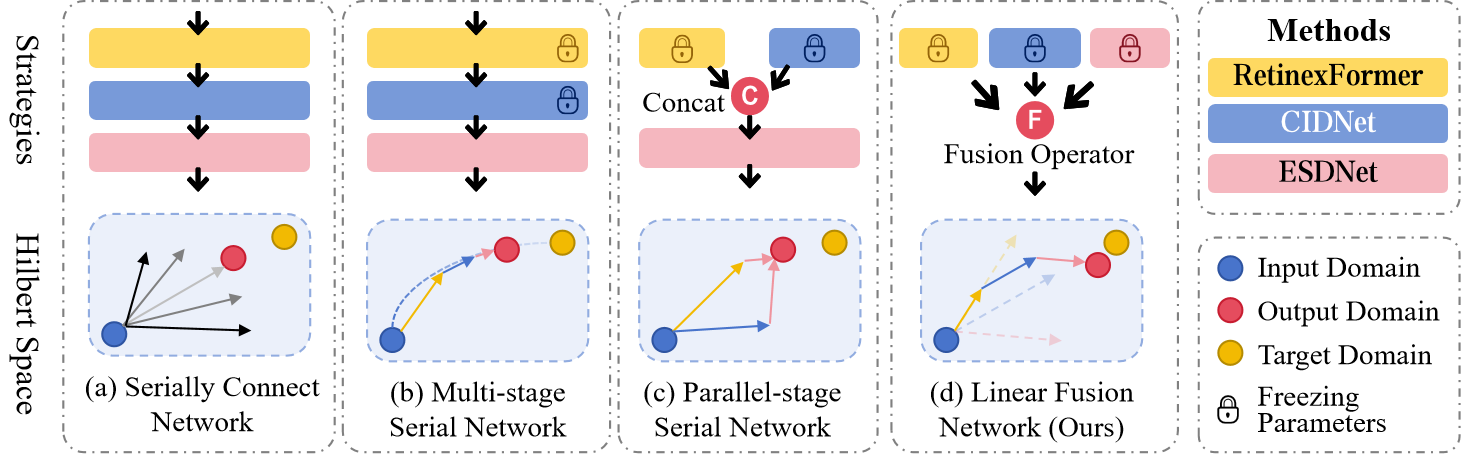}
    \caption{Four multi-model fusion strategies: (a) Serially Connect Network: Multiple networks are trained in a cascaded manner, but the probability of achieving optimal fitting in Hilbert space is low, making it difficult to converge to the best solution. (b) Multi-stage Serial Network: The parameters of the previous stage are frozen while training the next stage. However, this often leads to an increased number of iterations in later stages, making it more challenging to approach the optimal solution. (c) Parallel-stage Serial Network: Each model is trained separately in the earlier stages, and their outputs are concatenated before being fused by a new network. However, in Hilbert space, this method is fundamentally similar to (b) and does not resolve the issue of difficult convergence. (d) Linear Fusion Network: A simple yet efficient linear fusion operation is employed to enhance multi-model fusion performance, bringing the output space closer to the target space in Hilbert space.}
    \label{fig:motivation}
    \vspace{-0.15cm}
\end{figure*}
\section{Method}
\label{sec:Method}
\subsection{Overview}
Low-light image enhancement faces two major challenges. 
The first challenge is that the information on the content of the image is relatively scattered, requiring a large receptive field for enhancement tasks at this resolution. 
However, an excessively large receptive field may lead to the loss of fine-grained details. 
The second challenge is that enhancement tasks are typically conducted in the sRGB space, where color and brightness are often overfitted, leading to color distortion and artifacts.
Recently, the HVI color space has addressed the issues of sRGB. However, the parameter $k$ in the collapse function $\mathbf{C}_k$ needs to be trained specifically for a given dataset and method. 
When switching to a different dataset, changes in data distribution may render the learned collapse function ineffective, thus reducing generalization performance.

To address these two issues, we propose a hybrid network framework, FusionNet, which integrates three complementary approaches: ESDNet \cite{yu2022towards}, a CNN-based architecture for local feature processing; Retinexformer \cite{cai2023retinexformer}, a Transformer-based model capturing long-range dependencies; and CIDNet \cite{yan2025hvi}, which leverages the novel HVI color space. 
By combining the global receptive field of Transformers with the superior local feature extraction of CNNs, while simultaneously fusing the dataset-independent sRGB color space with the HVI space that decouples brightness from color, FusionNet enhances the generalization capability of image enhancement across diverse and unseen scenarios. 
Each of the three methods is individually trained and tested, and the final enhancement results are obtained by integrating their respective outputs. 

Next, we will present the specific fusion strategy and analyze why this integration leads to improved enhancement performance.

\subsection{Linear Fusion Method}
In FusionNet, the selection of the appropriate fusion strategy plays a crucial role. In the following sections, we first introduce four different fusion strategies and then explain the advantages of our proposed approach.
The first multi-network fusion strategy involves serially connecting multiple networks (as shown in Fig. \ref{fig:motivation}(a)) and performing single-stage training and testing on the resulting cascaded network. In some cases, pre-trained weights are also incorporated for overall fine-tuning to improve model fitting. 
However, this pipeline results in a deep network structure with a large number of parameters, which increases the risk of network collapse, where small parameter variations can cause significant performance fluctuations, making it difficult to achieve optimal results through single-stage training \cite{keskar2016large}.

The second strategy is a multi-stage serial network, which also connects different methods sequentially, as Fig. \ref{fig:motivation}(b). 
However, in each new training stage, the parameters of the previous stage are typically frozen. 
This approach mitigates the issue of deep network collapse but introduces new challenges: as training progresses, later stages often require more iterations, making the training process increasingly difficult \cite{pmlr-v38-lee15a}. 
Consequently, the time cost of multi-stage training significantly increases, and there is no guarantee that the later stages will converge to a more optimal solution.

The third strategy adopts a parallel training followed by serial enhancement approach (see Fig. \ref{fig:motivation}(c)). 
Specifically, multiple methods are first trained independently in a supervised manner, producing different outputs. 
These outputs are then concatenated and fed into a new network, which extracts features and fuses them into a final enhanced image, with the parameters of the parallel networks frozen. 
However, this approach, like the second strategy, still falls under the category of multi-stage training, failing to address the excessive time cost issue.

Finally, we propose a linear fusion strategy, which adopts a fully parallel execution structure where networks operate independently without mutual interference. 
In this approach, different methods are first trained separately in a supervised manner. Then, the final output is obtained by applying our linear fusion parameters to weight and sum the output images. This process can be formulated as follows:
\begin{equation}
\mathbf{I}_{HQ} = \sum_{i=1}^{n} k_i\mathtt{F}\left ( \mathbf{I}_{LQ}  \right ) ,
\label{eq:1}
\end{equation}
where $\mathbf{I}_{LQ}$ is a low-light image, $\mathbf{I}_{HQ}$ is the high quality output, and $\mathtt{F}(\cdot)$ represents the enhancement method of LLIE task. The parameter $k_i$ follows
\begin{equation}
    \sum_{i=1}^{n} k_i=a,
\end{equation}
where we set $a=1$ to ensure luminance stability of the fused image.
This fusion strategy ensures that different methods remain independent, allowing optimal results to be achieved solely by adjusting the final fusion parameters. 
Additionally, it eliminates the need for multi-stage training and incurs no extra time overhead, as multiple methods can be trained simultaneously. 
More importantly, this approach enables manual parameter tuning to significantly improve the final test performance and generalization capability. 
In the following section, we provide a mathematical explanation of the effectiveness of this approach.

\subsection{Hilbert Space Explanation}
As depicted in Fig. \ref{fig:motivation}, each enhancement method corresponds to a distinct mapping trajectory $f_i: \mathcal{X} \to \mathcal{Y}$ within Hilbert space \cite{rudin1955analytic}, where $\mathcal{X}$ represents the input domain and $\mathcal{Y}$ the target enhancement domain. 
The inner product defined in Eq. \ref{eq:1} establishes the quantitative relationship between enhancement operators and target characteristics:
\begin{equation}
    \langle f_i, f_t \rangle_{\mathcal{H}} = \int_{\mathcal{X}} f_i(x)f_t(x)d\mu(x),
\end{equation}
where the integral quantifies the expectation of correlation between enhancement operators over the input distribution $\mathcal{X}$, and $\mu(\cdot)$ is the Borel measure defining the integration over the input distribution.
This formulation measures the linear correlation between arbitrary enhancement method $f_i$ and the ideal target mapping $f_t$.
The optimal fusion of enhancement methods requires maximizing the projection magnitude in the target subspace:
\begin{equation}
    \max_{\alpha_i} \big\| \sum\nolimits_i \alpha_i \text{Proj}_{\mathcal{H}_t}(f_i) \big\|^2
\end{equation}
where the maximum projection condition is satisfied when $\{f_i\}$ maintain maximal linear independence in $\mathcal{H}_t$, as per the orthogonal projection theorem$^1$. 
\footnote{$^1$Orthogonal Projection:
for any closed subspace $\mathcal{H}_t \subset \mathcal{H}$ and $f \in \mathcal{H}$, the projection $\text{Proj}_{\mathcal{H}_t}(f)$ uniquely minimizes $\|f - g\|$ over all $g \in \mathcal{H}_t$.}

Therefore, we can think of the CNN-based (ESDNet), Transformer (Retinexformer), and HVI-based (CIDNet) methods used in FusionNet as near-orthogonal representations in reproducing kernel Hilbert space \cite{berlinet2011reproducing} enable effective target domain coverage through convex combination as Eq. \ref{eq:1}, improving the generalization performance of our method.

\subsection{Training Loss Functions}
We follow the loss functions in the original paper of ESDNet \cite{yu2022towards}, RetinexFormer \cite{cai2023retinexformer}, and CIDNet \cite{yan2025hvi}, which combines pixel-level and feature-level supervision for effective training. Each of the three methods is individually trained using its respective loss function, and their outputs are subsequently fused during the testing phase.

\section{Experiment}
\subsection{Experimental Settings and Details}

We conduct comprehensive training and evaluation on both the NTIRE2025 Low Light Enhancement Challenge dataset \cite{liu2025ntire} and the LOL dataset.

\noindent\textbf{NTIRE2025.} It is a Ultra-High-Definition (UHD) collection featuring images with resolutions up to 4K and beyond, consisting of 219 training scenes, 46 validation scenes, and 30 testing scenes. During the competition, our training pipeline was carefully designed to maximize model performance in LLIE. Specifically, we cropped the training images into $1024\times1024$ patches and trained the CIDNet model for 90 kilo-iterations using a batch size of 1. For the ESDNet model, we employed larger patches of $1600\times1600$ and trained for 100 kilo-iterations with a batch size of 1. Similarly, the Retinexformer model was trained on $ 2000\times2000$ patches for 180 kilo-iterations, also with a batch size of 1. This tailored approach allowed us to optimize the convergence and efficacy of each model within the competitive framework of the NTIRE2025 Low Light Enhancement Challenge.

\noindent\textbf{LOL.} The LOL dataset comprises two versions, LOLv1 \cite{wei1808deep} and LOLv2 \cite{lol_v2}. LOLv2 dataset further divided into real and synthetic subsets. The data is partitioned with training-to-test ratios of 485:15 for LOLv1, 689:100 for LOLv2-real, and 900:100 for LOLv2-synthetic. For training, we cropped the images into 384 × 384 patches and trained three distinct models for 200 kilo-iterations using a batch size of 16, ensuring robust convergence and effective performance across varying low-light conditions.

\noindent\textbf{Evaluation Metrics.} We evaluate distortion using Peak Signal-to-Noise Ratio (PSNR) and Structural Similarity (SSIM) \cite{SSIM}. In addition, to assess the perceptual quality of restored images, we employ the Learned Perceptual Image Patch Similarity (LPIPS) metric \cite{LPIPS}, using AlexNet \cite{Alex} as the reference network.

\noindent\textbf{Random Gamma Curve.}
We observed that employing the random gamma curve technique ($\mathbf{I_{Input}}=\mathbf{I_{LQ}}^\gamma$, where $\gamma$ is the random number between 0.6 and 1.2) for data pre-processing enhances the generalization capability of the model. 
To improve performance on the NTIRE challenge dataset, we use this technique on CIDNet for data augmentation.

\noindent\textbf{Implementation details.} 
All models are implemented in PyTorch with Adam optimizer ($\beta_1 = 0.9$ and $\beta_2 = 0.999$) and cosine annealing scheme \cite{sgdr}. All the training configuration details are given in Tab. \ref{tab:training_config}.

\begin{table}
\centering
\caption{Training configurations for different models. LR represents learning rate in training process.}
\label{tab:training_config}
\resizebox{\linewidth}{!}{
\begin{tabular}{c|c|c}
\toprule
\textbf{Model} & \textbf{Hardware}  & \textbf{Initial/Final LR} \\
\midrule
RetinexFormer \cite{cai2023retinexformer} & Tesla A100  & $2\times10^{-4} \rightarrow 1\times10^{-6}$ \\
ESDNet \cite{yu2022towards} & Tesla A100  & $2\times10^{-4} \rightarrow 1\times10^{-6}$  \\
CIDNet \cite{yan2025hvi}& RTX 4090  & $1\times10^{-4} \rightarrow 1\times10^{-7}$ \\
\bottomrule
\end{tabular}}
\vspace{-0.22cm}
\end{table}
\begin{table*}[ht]
    \centering
    \small
    \caption{The final ranking score of top-five teams' results of the NTIRE2025 Low Light Enhancement Challenge. 
    The final ranking is determined based on a weighted sum of different ranking criteria: Rank P 0.5 (weight: 0.5), Rank S 0.5 (weight: 0.5), Rank L 0.4 (weight: 0.4), and Rank N 0.2 (weight: 0.2).}
    \resizebox{0.85\linewidth}{!}{
    \begin{tabular}{l|cccc|cccc|c}
        \toprule
        Team &  PSNR$\uparrow$ & SSIM$\uparrow$ & LPIPS$\downarrow$ & NIQE$\downarrow$ & Rank P& Rank S& Rank L & Rank N& Total Rank$\downarrow$ \\
        \midrule
        NWPU-HVI (Ours) &  26.24  & 0.861 & 0.128 & 10.9539 & 2  & 2  & 7  & 11 & 7    \\
        Imagine  &  26.345 & 0.858 & 0.133 & 11.8073 & 1  & 3  & 9  & 23 & 10.2  \\
        pengpeng-yu & 25.849 & 0.858 & 0.134 & 11.2933 & 4  & 3  & 11  & 16 & 11.1  \\
        DAVIS-K  & 25.138 & 0.863 & 0.127 & 10.5814 & 14 & 1  & 6  & 9  & 11.7  \\
        SoloMan  & 25.801 & 0.856 & 0.130 & 11.4979 & 5  & 6  & 8  & 19 & 12.5  \\
        \bottomrule
    \end{tabular}}
    \label{tab:NTIRE2025}
\end{table*}
\begin{table*}[t]
\centering
{\caption{Quantitative comparisons of different methods on LOLv1 and LOLv2. The best and second best results are highlighted in \textbf{bold} and \underline{underlined}, respectively. Note that we obtained these results either from the corresponding papers, or by running the pre-trained models released by the authors, and some of them lack relevant results.}
\resizebox{0.85\textwidth}{!}{
\begin{tabular}{l|c|c|ccc|ccc|ccc}
\toprule
\multirow{2}{*}{\textbf{Methods}} 
&\multirow{2}{*}{\textbf{Params (M)}}
&\multirow{2}{*}{\textbf{FLOPs (G)}} 
& \multicolumn{3}{c|}{\textbf{LOLv1}} 
& \multicolumn{3}{c|}{\textbf{LOLv2-real}}  
& \multicolumn{3}{c}{\textbf{LOLv2-synthetic}}   \\  
& & &PSNR $\uparrow$   & SSIM $\uparrow$  & LPIPS $\downarrow$ & PSNR $\uparrow$   & SSIM $\uparrow$   & LPIPS $\downarrow$ & PSNR $\uparrow$   & SSIM $\uparrow$   & LPIPS $\downarrow$ \\ \midrule

LIME \cite{guo2016lime}        &-            &-                  & 16.76  & 0.560 & 0.350 & 15.24  & 0.470  & 0.415 & 16.88  & 0.776  & 0.675 \\
RetinexNet \cite{liu2021retinex}     &0.84  &584.47                  & 16.77  & 0.419 & 0.474 & 16.09  & 0.401  & 0.365 & 17.13  & 0.762  & 0.754 \\
EnlightenGAN \cite{jiang2021enlightengan}      &114.35        &61.01                  & 17.48  & 0.652 & 0.322 & 18.64  & 0.677  & 0.309 & 16.57  & 0.734  & -     \\
LLFlow \cite{wang2022low}         &17.42     &358.4                  & 21.14  & 0.854 & 0.126 & 17.43  & 0.831  & 0.149 & 24.80  & 0.919  & 0.074 \\
SNR-Aware \cite{xu2022snr}            &4.01    &26.35                  & 24.61  & 0.842 & 0.163 & 21.48  & 0.849  & 0.169 & 24.14  & 0.928  & 0.065 \\
MIRNet \cite{mirnet}             &31.76    &785                  & 24.14  & 0.830 & -     & 20.02  & 0.820  & -     &21.94  & 0.876  & -     \\
LLFormer \cite{wang2023ultra}       &24.55   &22.52            & 23.64  & 0.816 & 0.174 & 20.05  & 0.792  & 0.213 & 24.03  & 0.909  & 0.064 \\
PyDiff \cite{zhou2023pyramid}         &97           &-                  & 22.59  & 0.851 & 0.110 & -      & -      & -     & -      & -      & -     \\
FourLLIE \cite{wang2023fourllie}       &0.12    &1.95                  & -      & -     & -     & 22.34  & 0.846  & 0.159 & 24.65  & 0.919  & 0.068 \\
RetinexFormer \cite{cai2023retinexformer}   &1.53  &15.85                  & \underline{25.15}  & 0.845 & 0.131 & 21.23  & 0.838  & 0.171 & 25.67  & 0.930  &  0.061 \\
GSAD \cite{hou2024global}         &17.36        &442.02                  & 22.56  & 0.849 & \underline{0.104} & 20.15  & 0.845  & \underline{0.113} & 24.47  & 0.928  & 0.063 \\
QuadPrior \cite{wang2024zero}  &1252.75 &1103.20 & 	22.85& 	0.800  & 0.201&	20.59  &	0.811  & 0.202&	16.11 &	0.758 &	0.114
\\
RetinexMamba \cite{bai2024retinexmamba}     &-     &42.82                   & 24.02  & 0.827 & 0.146 & 22.45  & 0.844  & 0.174 & \underline{25.88}  & 0.935  & 0.057 \\
CIDNet \cite{yan2025hvi}        &1.88             &7.57                  & 23.50  & \textbf{0.870} & 0.105 & \underline{23.90}  & \underline{0.871}  & 0.118 & 25.70  & \underline{0.942}  & \underline{0.047} \\
Ours             &13.54          &50.29                  & \textbf{25.17}  & \underline{0.857} & \textbf{0.103} & \textbf{24.44}  & \textbf{0.875}  & \textbf{0.110} & \textbf{26.50}  & \textbf{0.945}  & \textbf{0.042} \\ \bottomrule
\end{tabular}}}
\end{table*}
\label{tab:lolv12}
\subsection{Quantitative Evaluations} 
\textbf{Results on NTIRE Dataset.} 
Tab. \ref{tab:NTIRE2025} presents a comparison of our FusionNet method with the top five teams in the NTIRE competition. Ultimately, we achieved a decisive lead, surpassing the second-place team by 3.2 ranking positions overall.
Specifically, our method ranked second in both PSNR and SSIM, while LPIPS and NIQE were not within the top five. 
However, since these two perceptual quality metrics do not directly assess pixel-wise accuracy, the competition organizers assigned them relatively lower weights. Notably, we observed that the NIQE metric showed almost no correlation with PSNR or the final rankings. 
In some cases, even severely overexposed images exhibited better NIQE scores, suggesting that this metric may not be particularly reliable for low-light image enhancement.

\noindent\textbf{Results on LOL Datasets.} 
To further validate the effectiveness of our fusion method across more diverse datasets, we conducted additional experiments on two subsets of LOLv1 and LOLv2. 
As shown in Tab. \ref{tab:lolv12}, our approach outperformed 14 SOTA methods across eight evaluation metrics, with the only exception being the SSIM score on LOLv1, where it was slightly lower than CIDNet.
This may be attributed to the inclusion of ESDNet, a lightweight CNN-based model, in our method. While ESDNet is efficient, it has weaker noise suppression capabilities. 
Since LOLv1 contains a substantial amount of noise caused by low-light conditions, ESDNet struggled to remove such noise effectively, ultimately introducing a negative impact within FusionNet. 
In contrast, LOLv2-syn primarily exhibits color degradation due to low-light conditions, with minimal noise interference. 
In this scenario, ESDNet, designed for moiré pattern removal, proves to be more effective. Consequently, our method achieved a significant improvement in PSNR, surpassing RetinexMamba by 0.62 dB, further demonstrating that linear fusion enhances the model’s generalization capability.

\begin{figure*}[!t]
\centering
\begin{minipage}[t]{0.12\linewidth}
    \centering
    \vspace{3pt}
    \centering{\includegraphics[width=\textwidth]{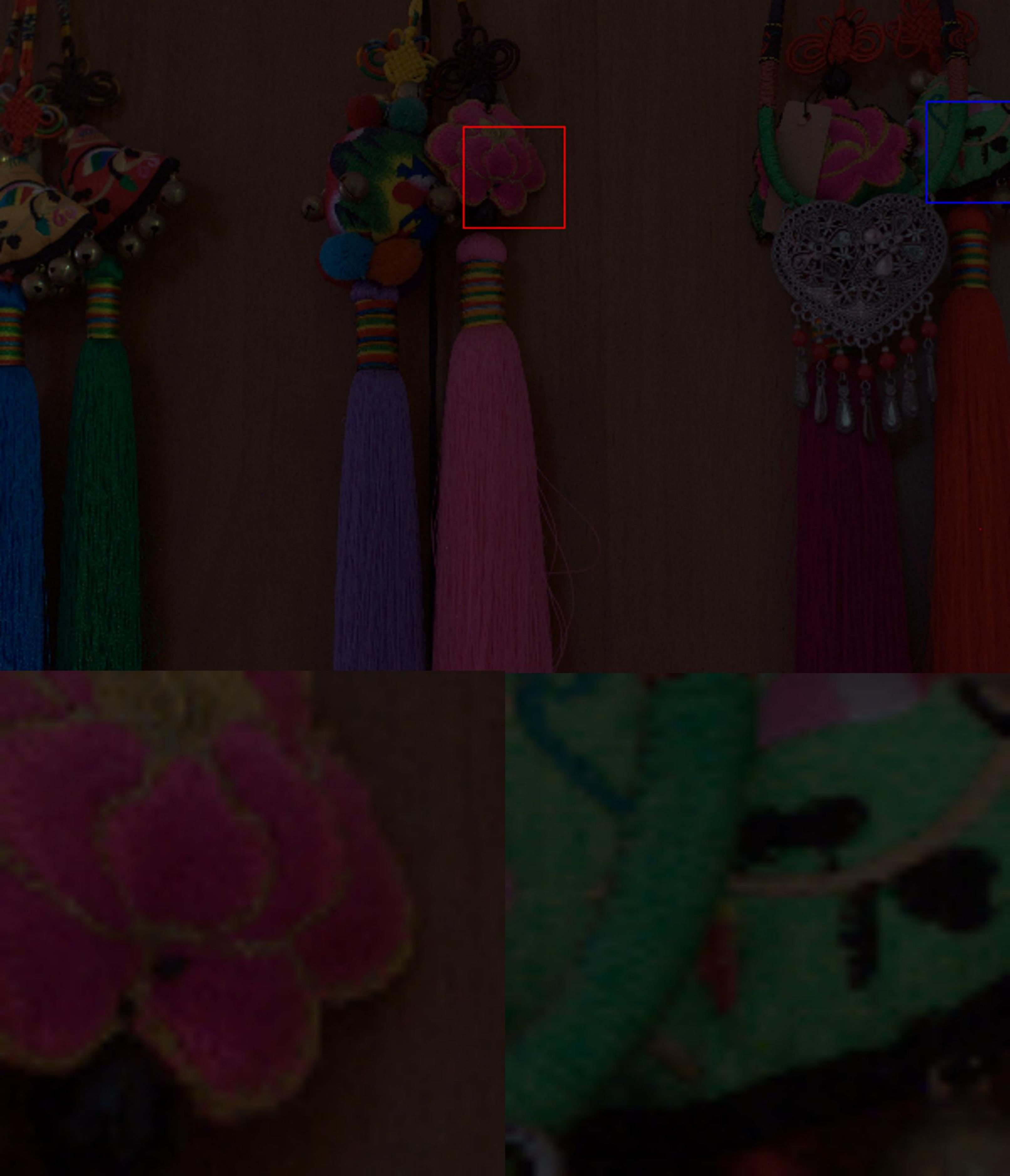}
    \centering{\includegraphics[width=\textwidth]{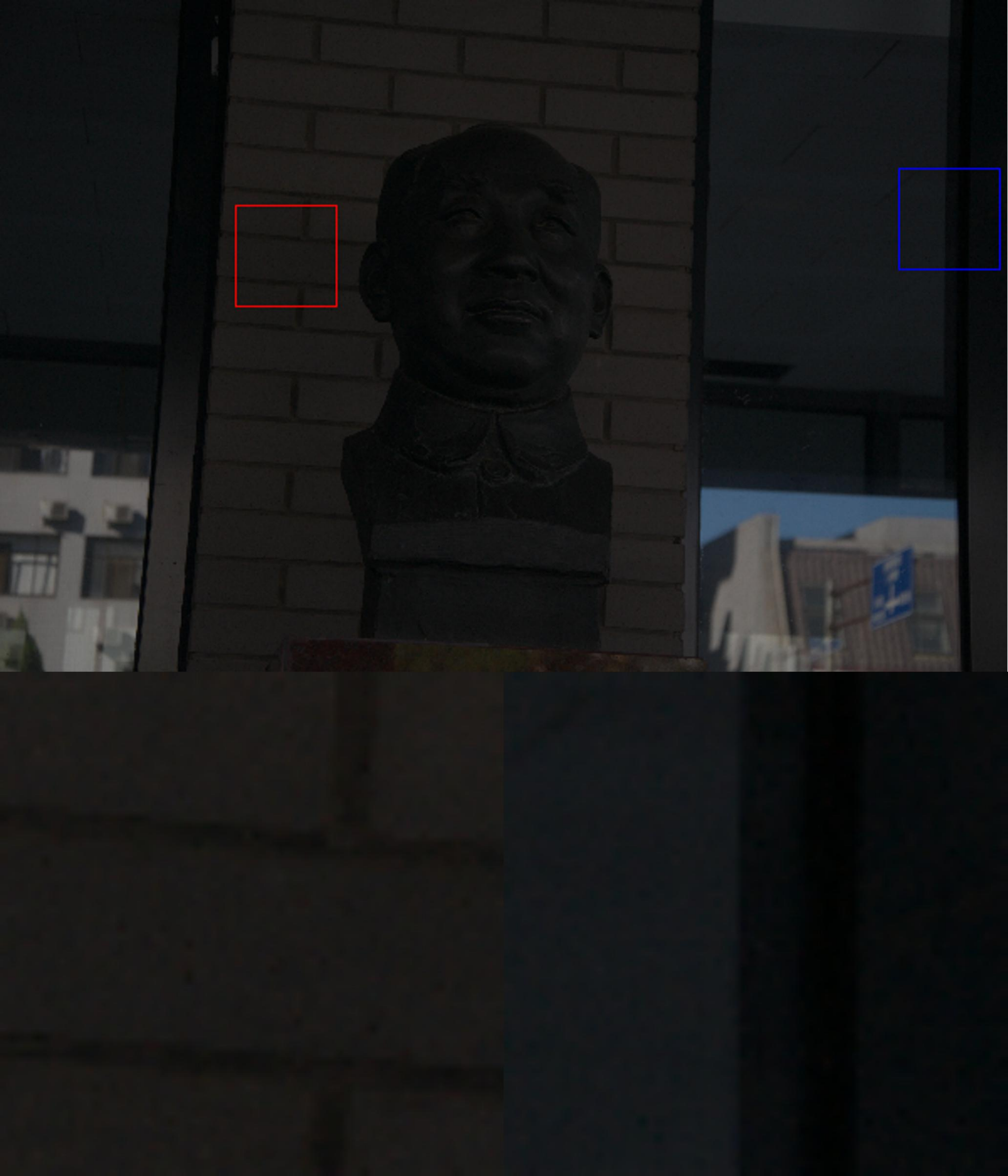}}
    \centering{\includegraphics[width=\textwidth]{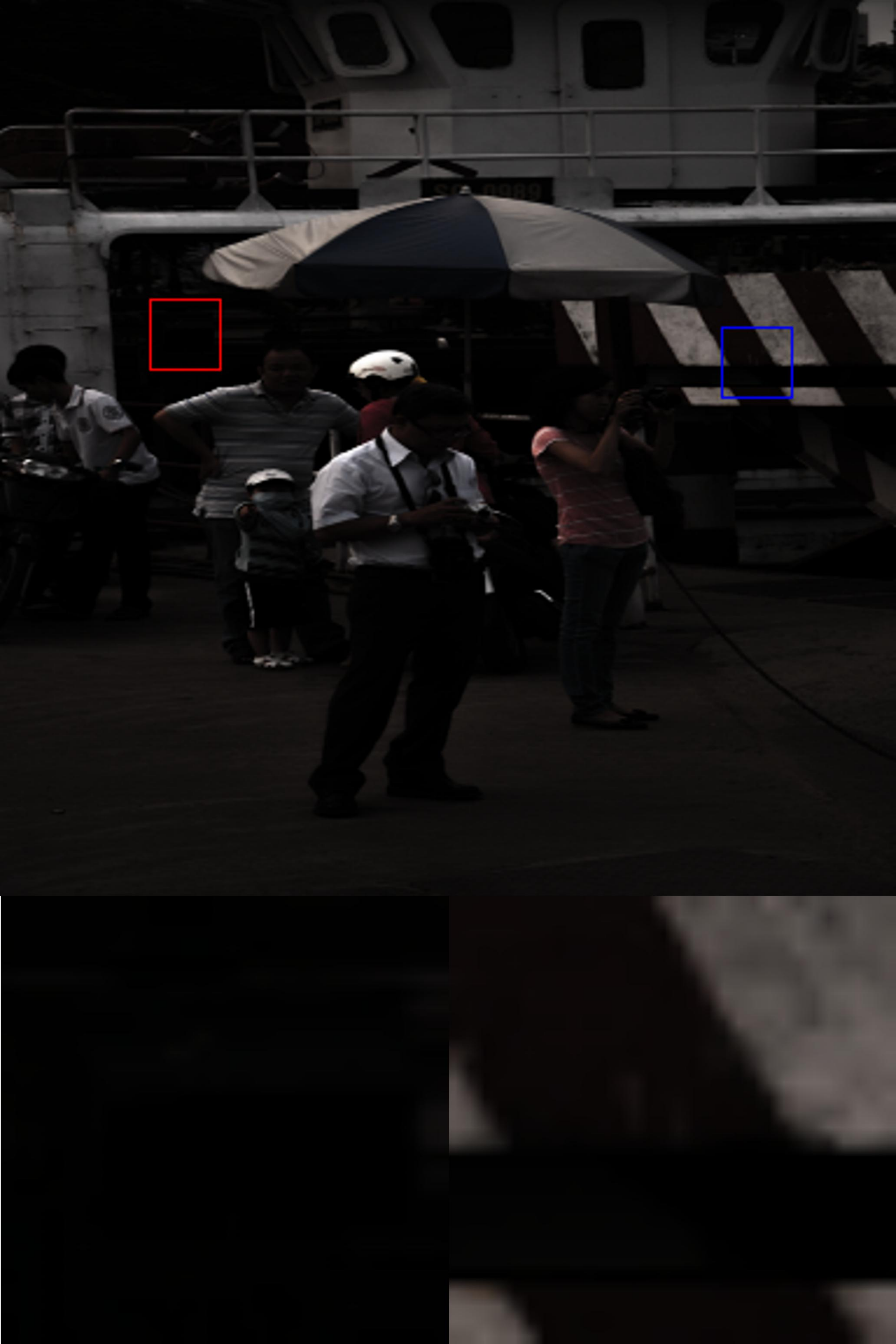}}
    \centerline{\small (a) Input}}
\end{minipage}
\begin{minipage}[t]{0.12\linewidth}
    \centering
    \vspace{3pt}
    \centering{\includegraphics[width=\textwidth]{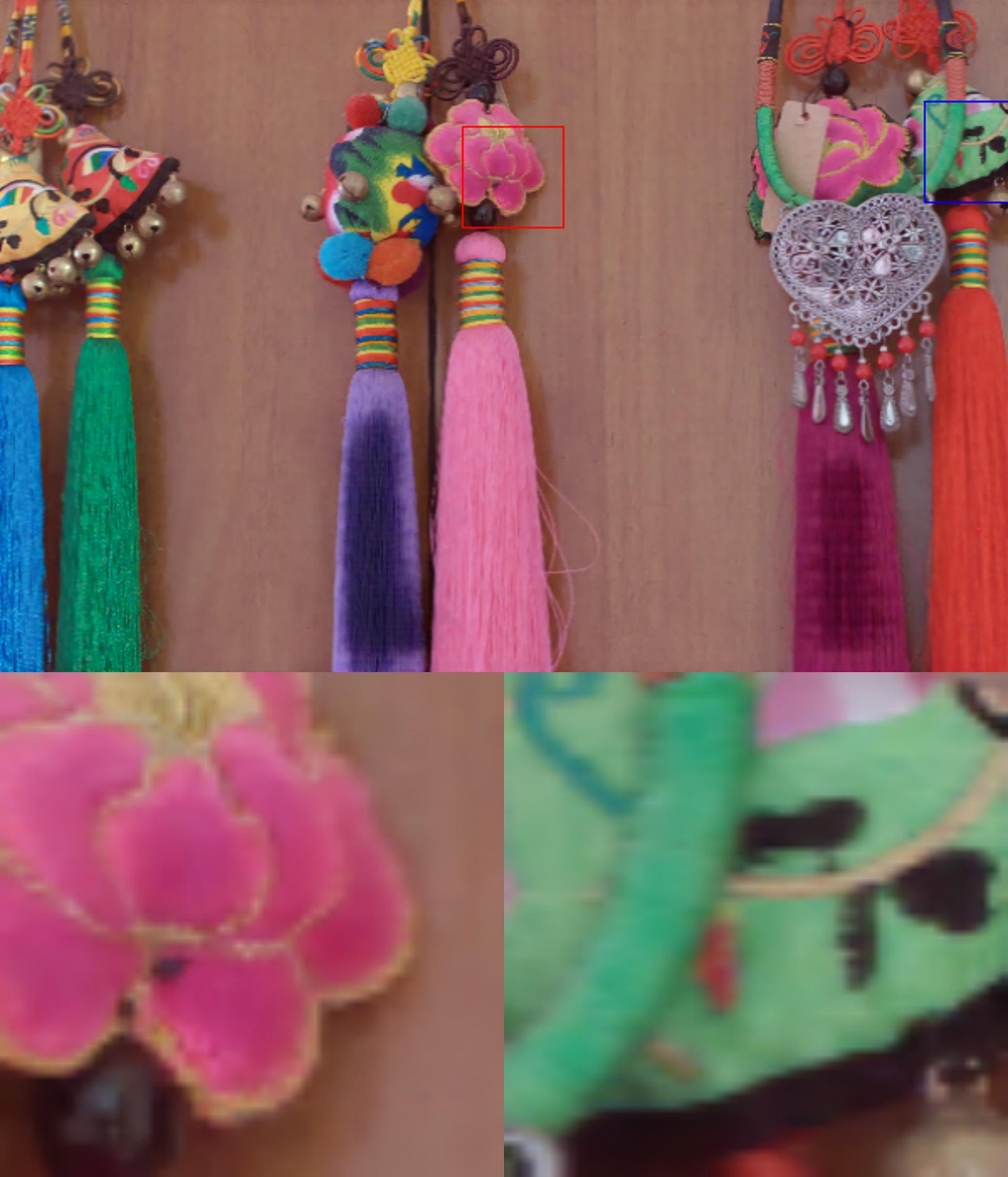}}
    \centering{\includegraphics[width=\textwidth]{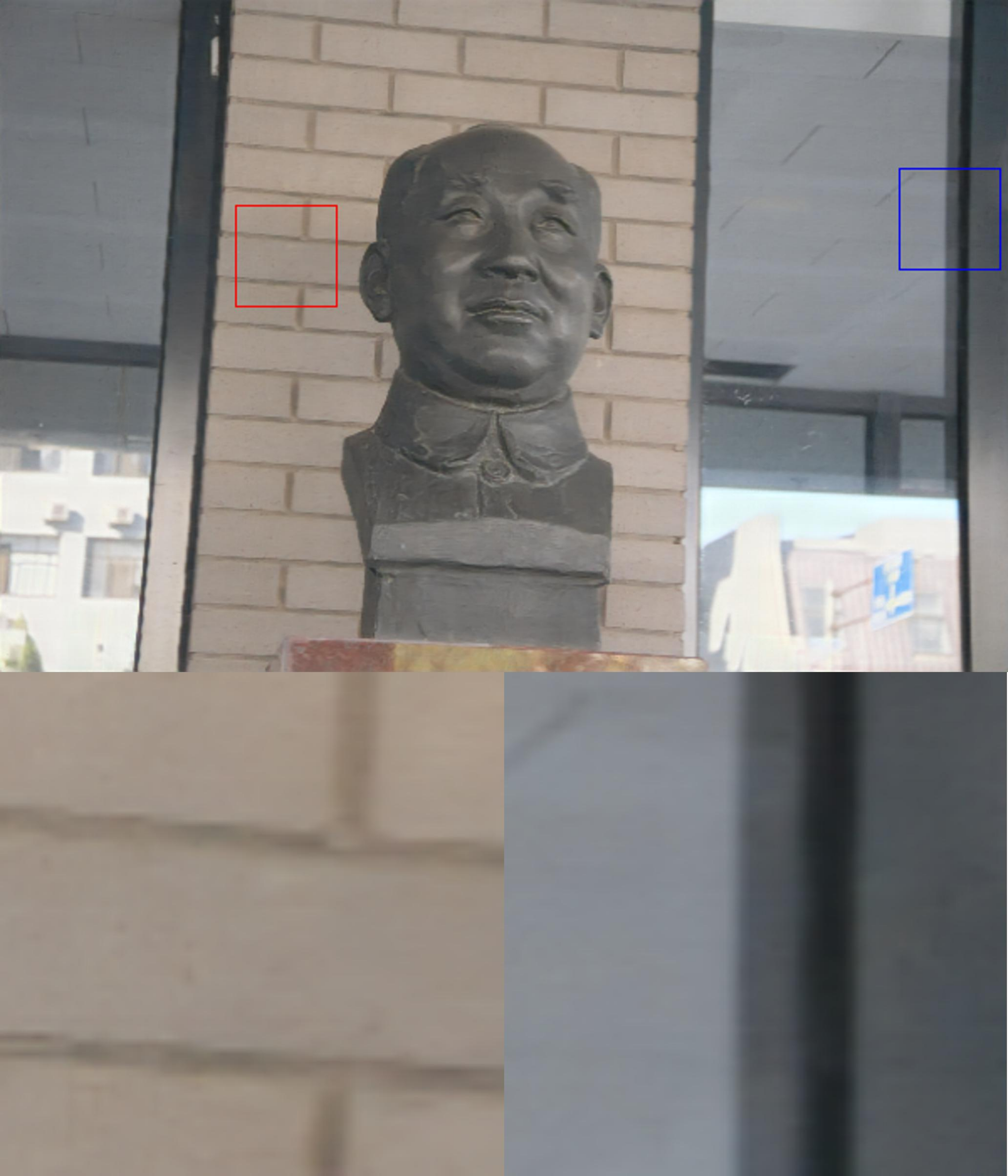}}
    \centering{\includegraphics[width=\textwidth]{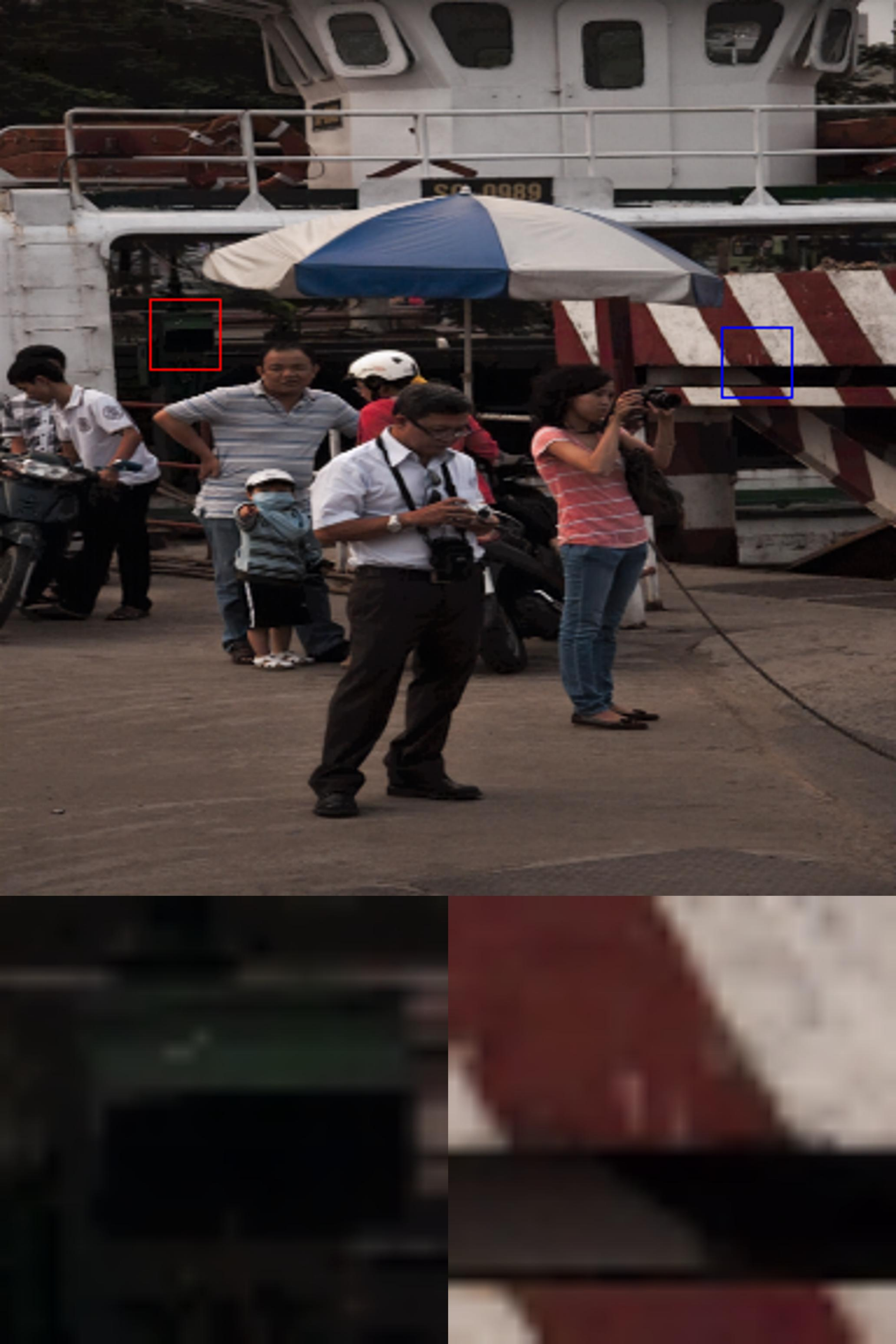}}
    \centerline{\small (b) SNRNet}
\end{minipage}
\begin{minipage}[t]{0.12\linewidth}
    \centering
    \vspace{3pt}
    \centering{\includegraphics[width=\textwidth]{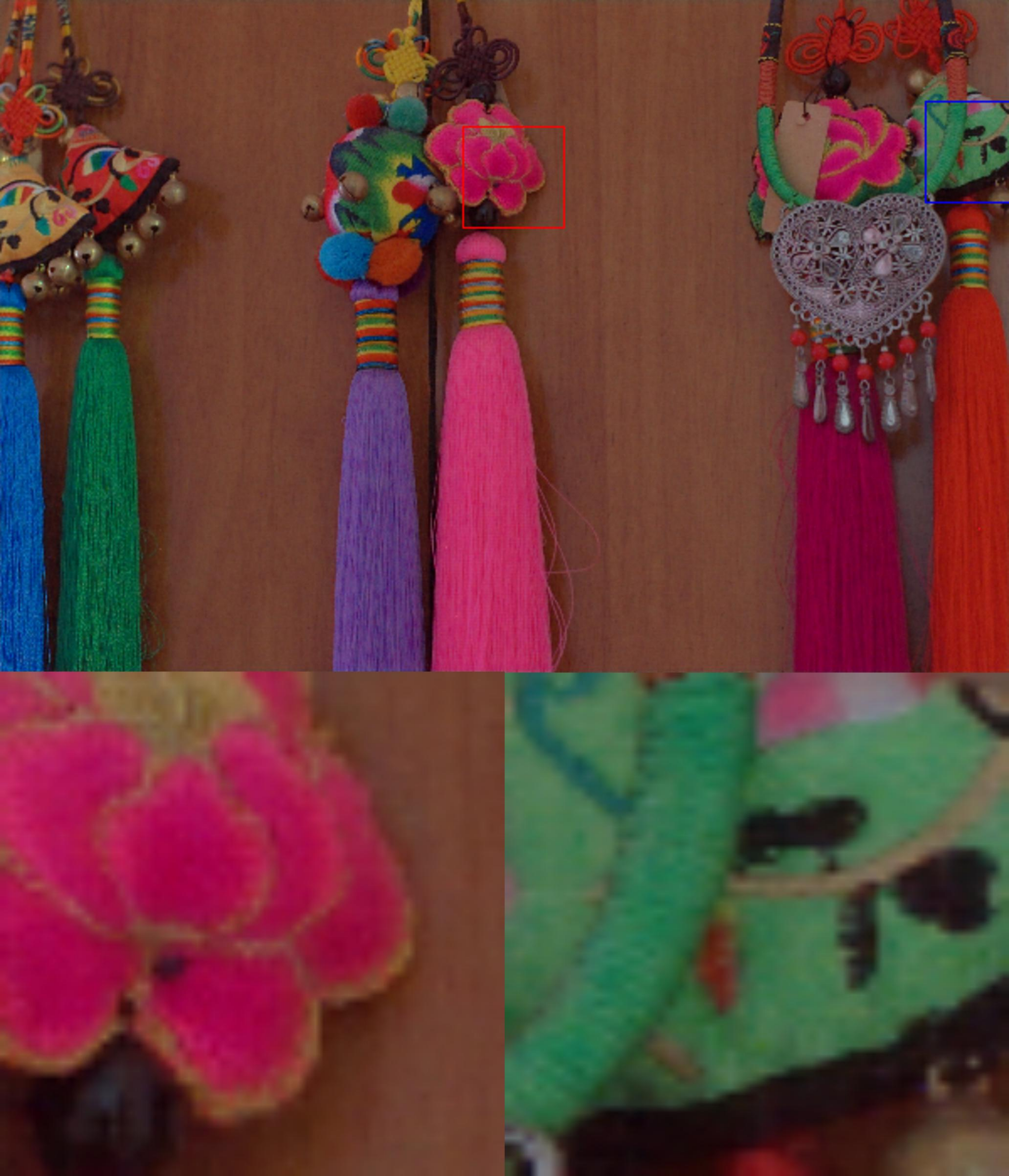}}
    \centering{\includegraphics[width=\textwidth]{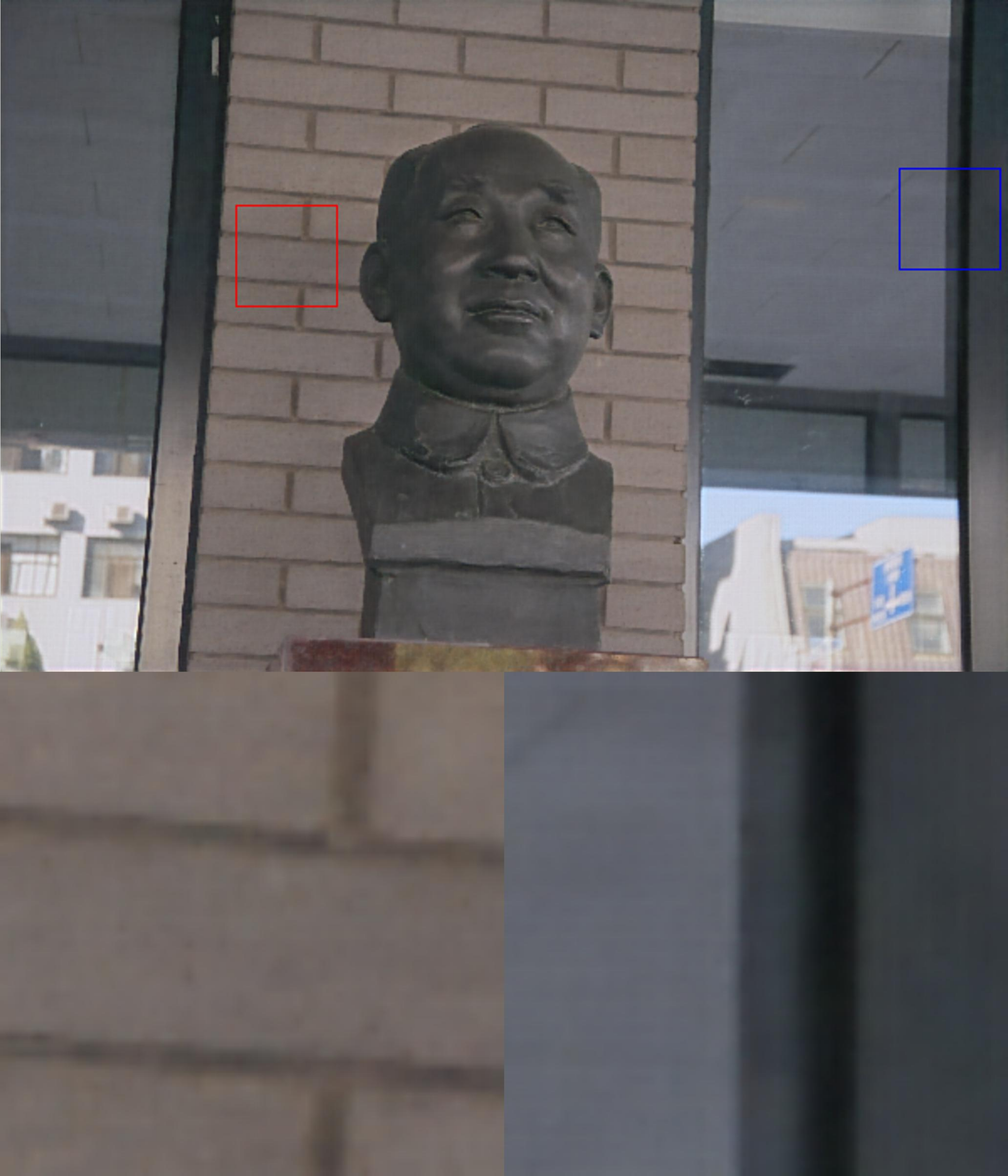}}
    \centering{\includegraphics[width=\textwidth]{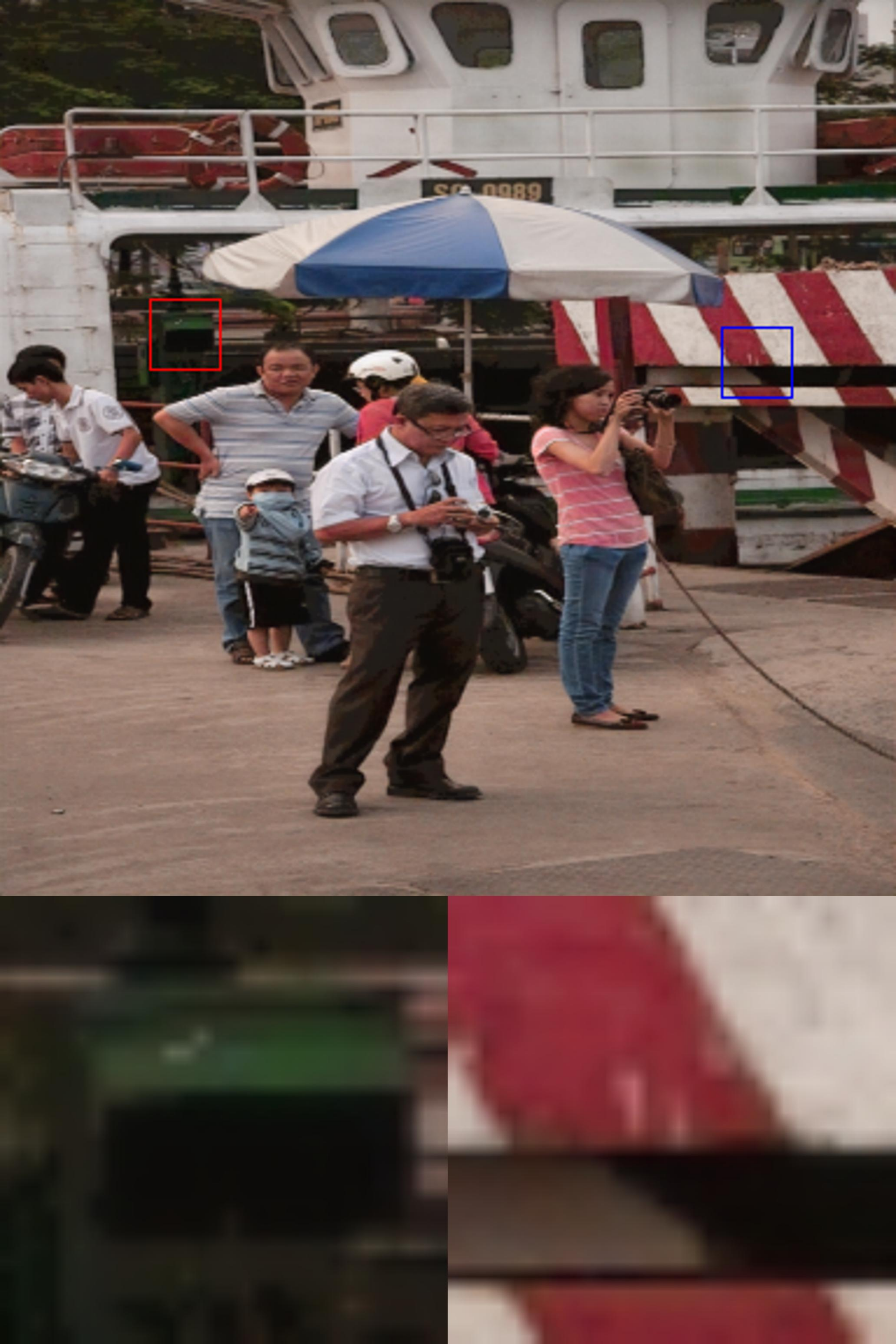}}
    \centerline{\small (c) RetFormer}
\end{minipage}
\begin{minipage}[t]{0.12\linewidth}
    \centering
    \vspace{3pt}
    \centering{\includegraphics[width=\textwidth]{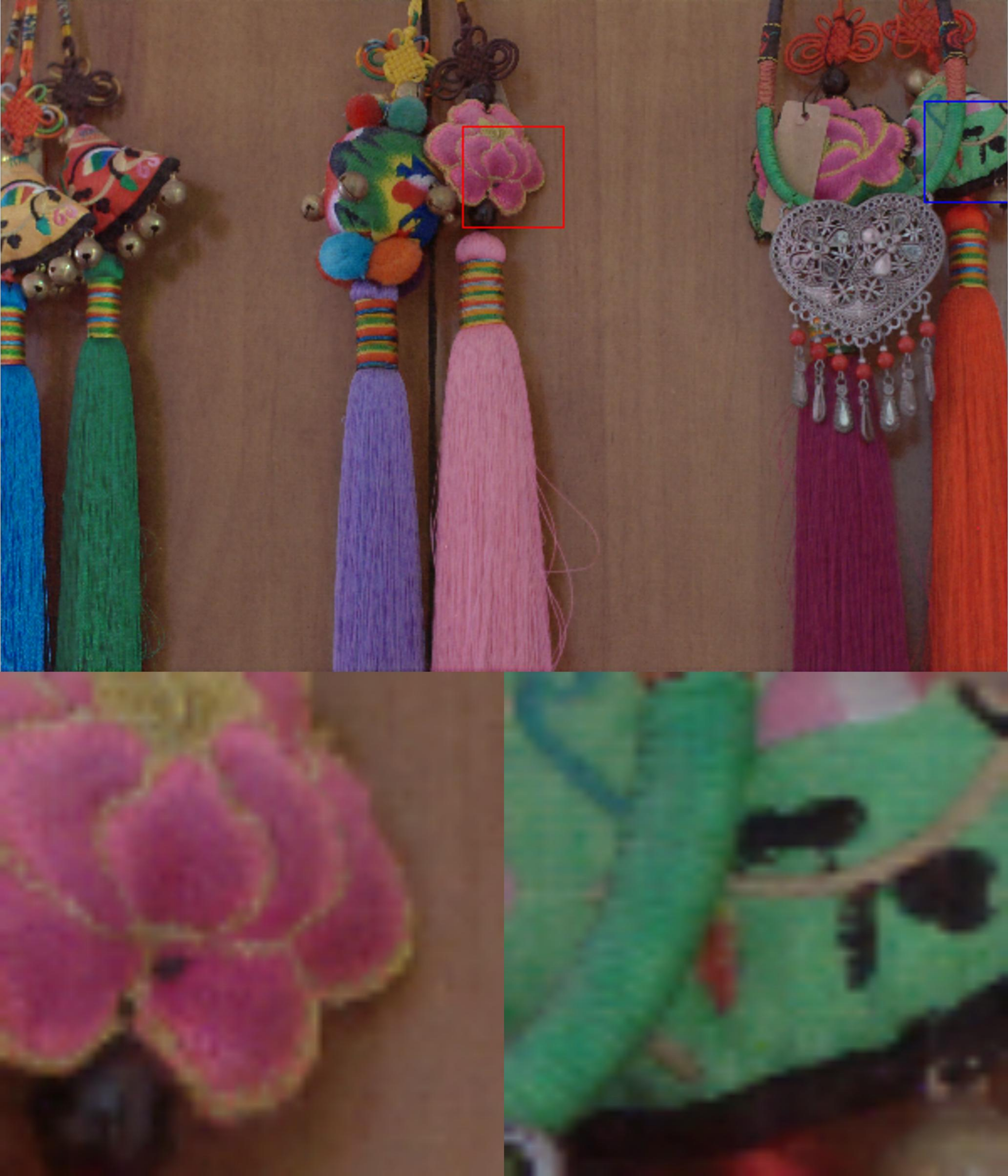}}
    \centering{\includegraphics[width=\textwidth]{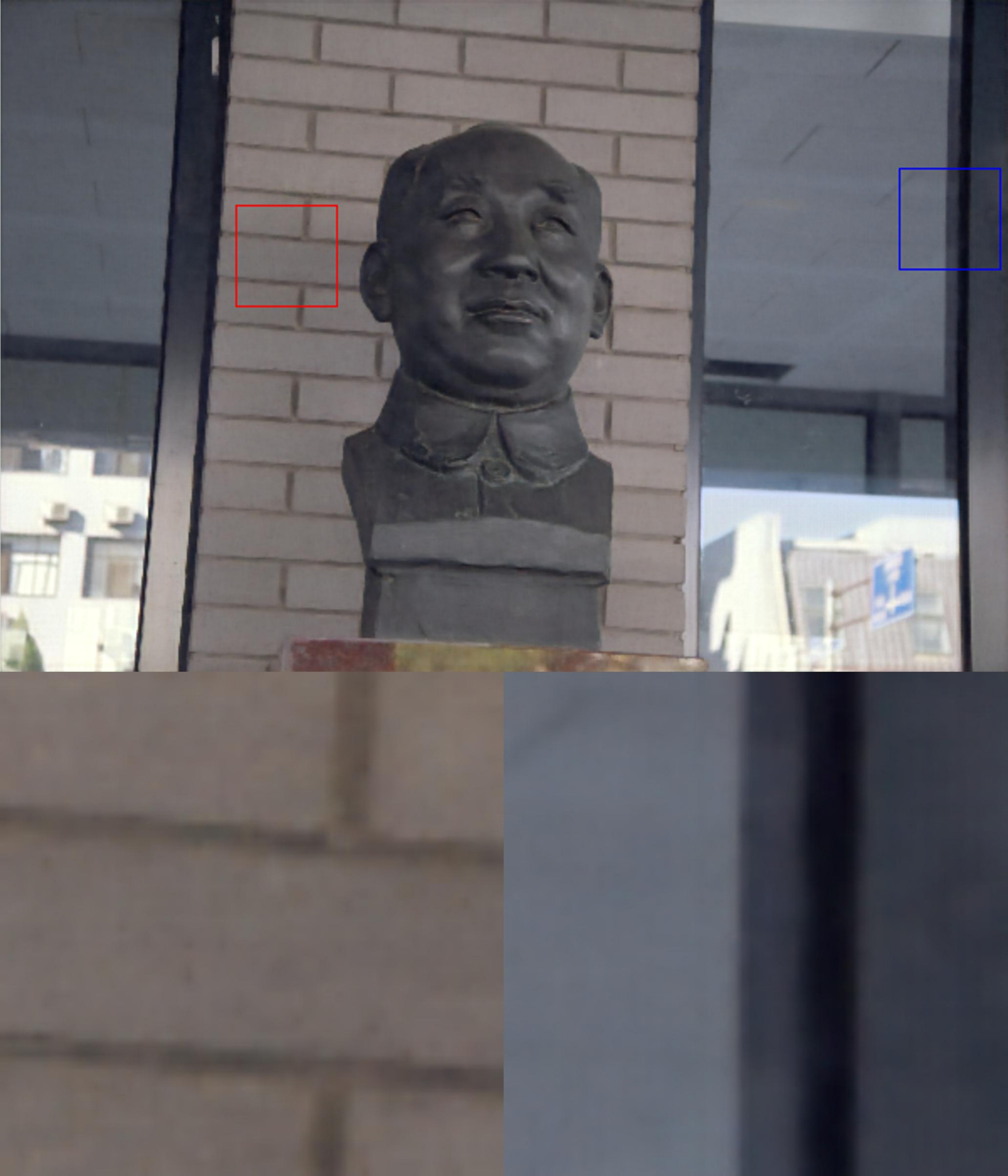}}
    \centering{\includegraphics[width=\textwidth]{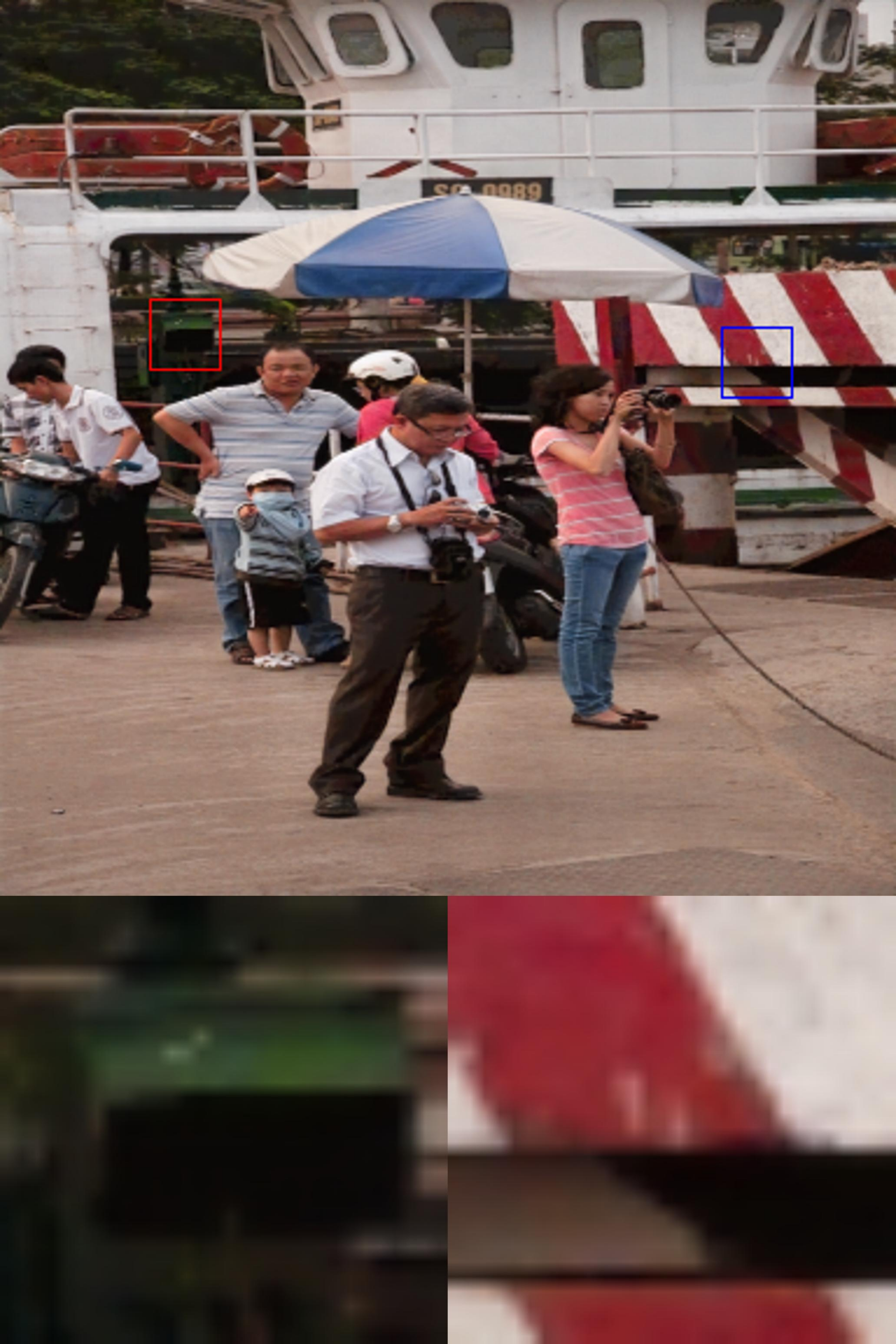}}
    \centerline{\small (d) RetMamba}
\end{minipage}
\begin{minipage}[t]{0.12\linewidth}
    \centering
    \vspace{3pt}
    \centering{\includegraphics[width=\textwidth]{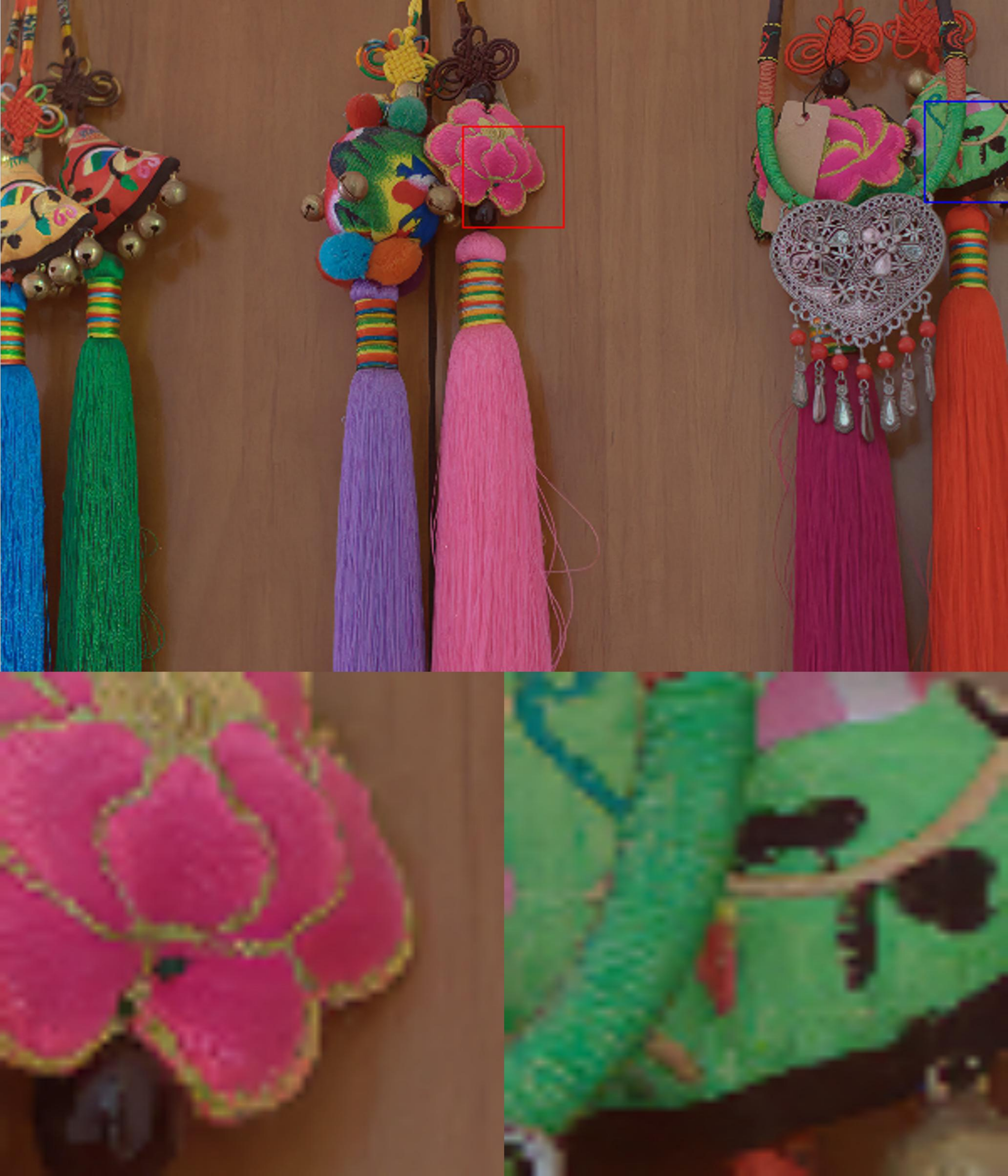}}
    \centering{\includegraphics[width=\textwidth]{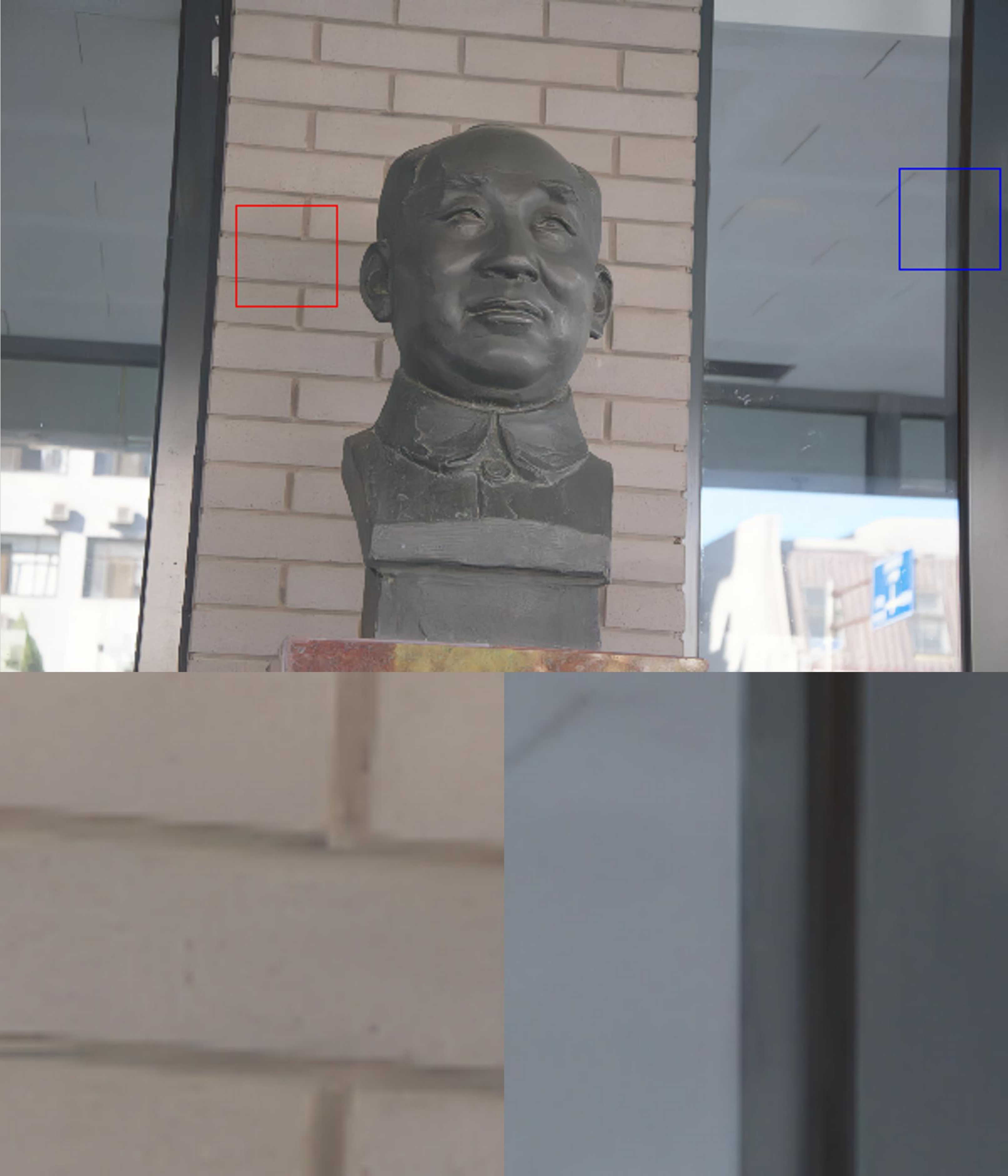}}
    \centering{\includegraphics[width=\textwidth]{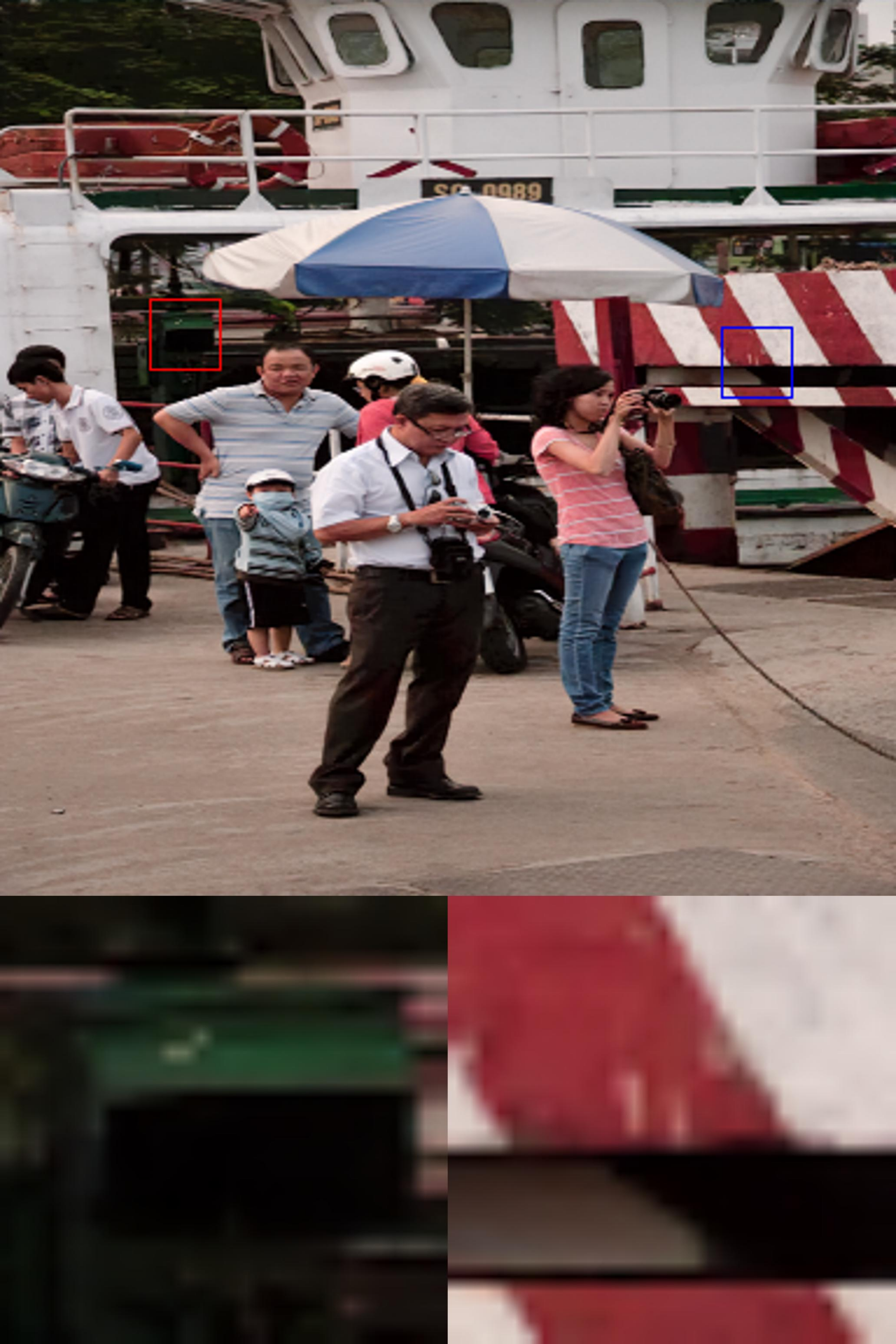}}
    \centerline{\small (e) GSAD}
\end{minipage}
\begin{minipage}[t]{0.12\linewidth}
    \centering
    \vspace{3pt}
    \centering{\includegraphics[width=\textwidth]{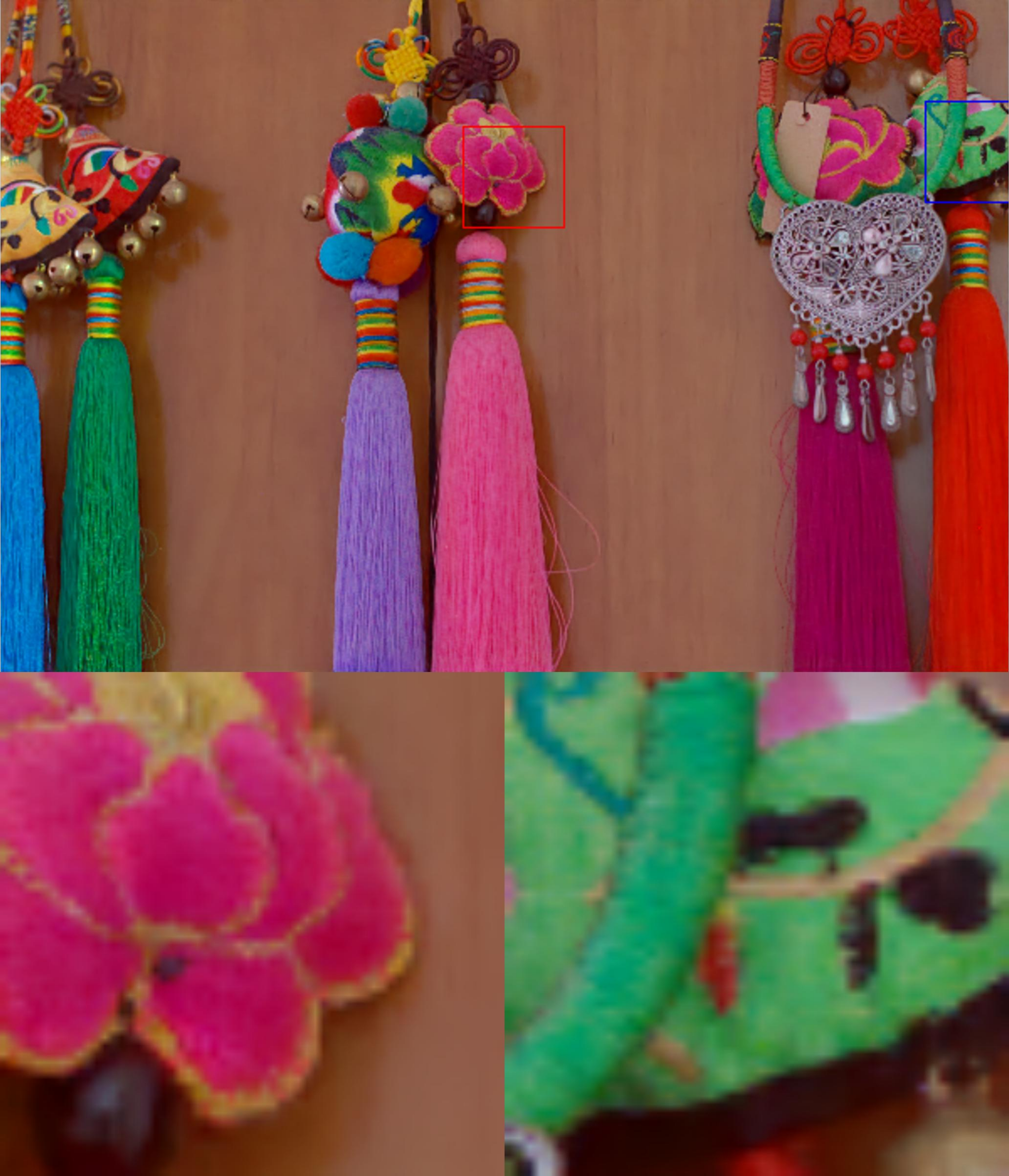}}
    \centering{\includegraphics[width=\textwidth]{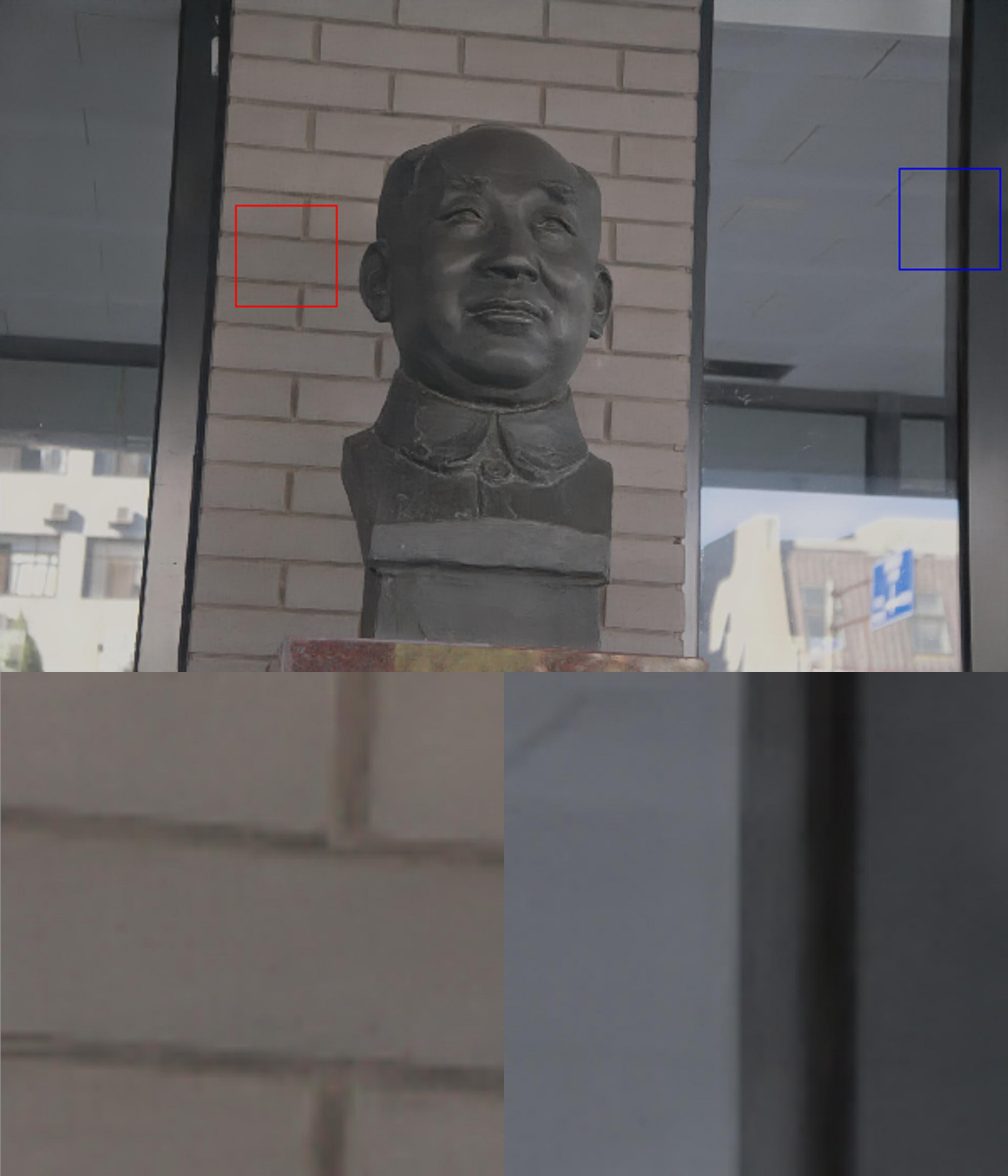}}
    \centering{\includegraphics[width=\textwidth]{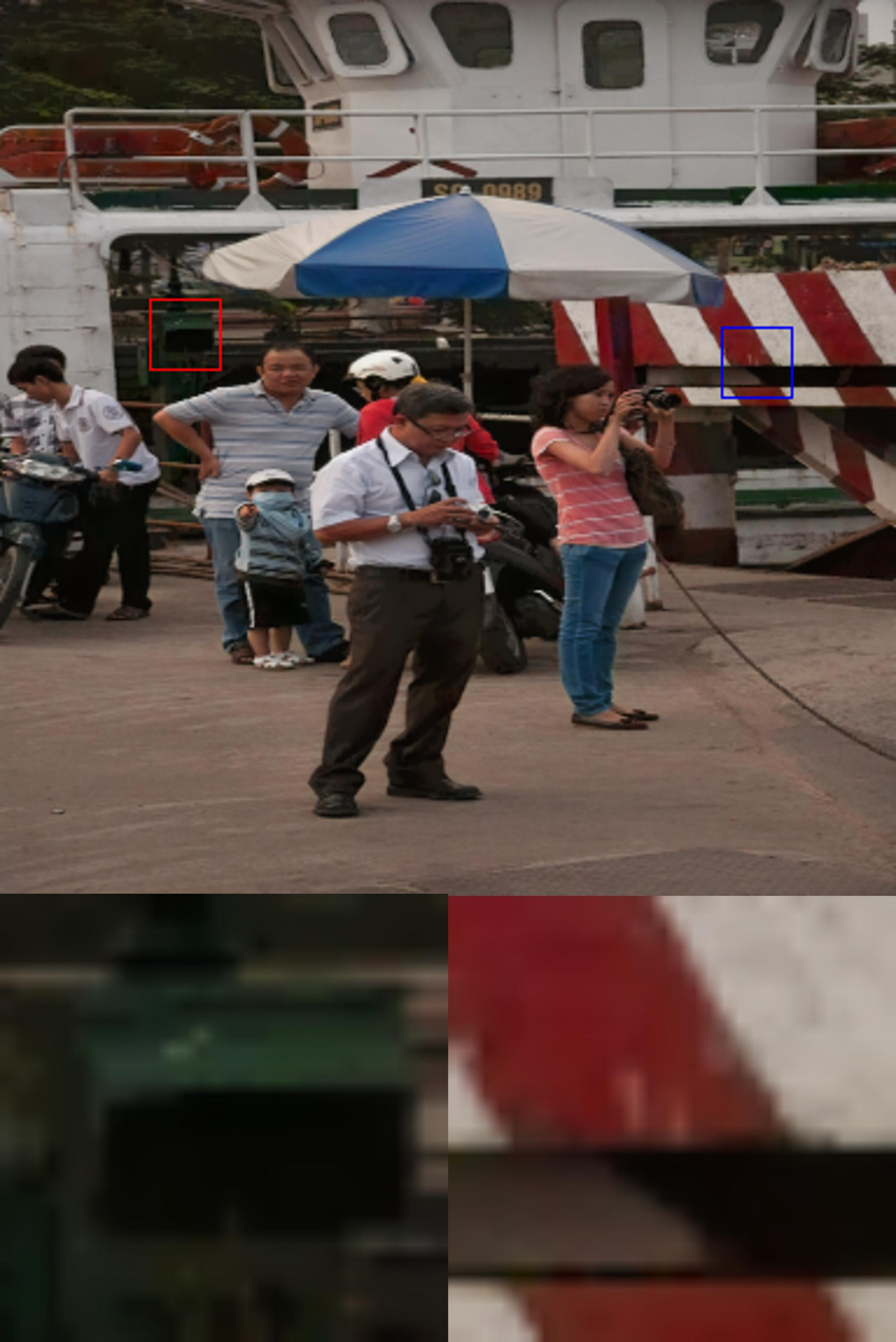}}
    \centerline{\small (f) CIDNet}
\end{minipage}
\begin{minipage}[t]{0.12\linewidth}
    \centering
    \vspace{3pt}
    \centering{\includegraphics[width=\textwidth]{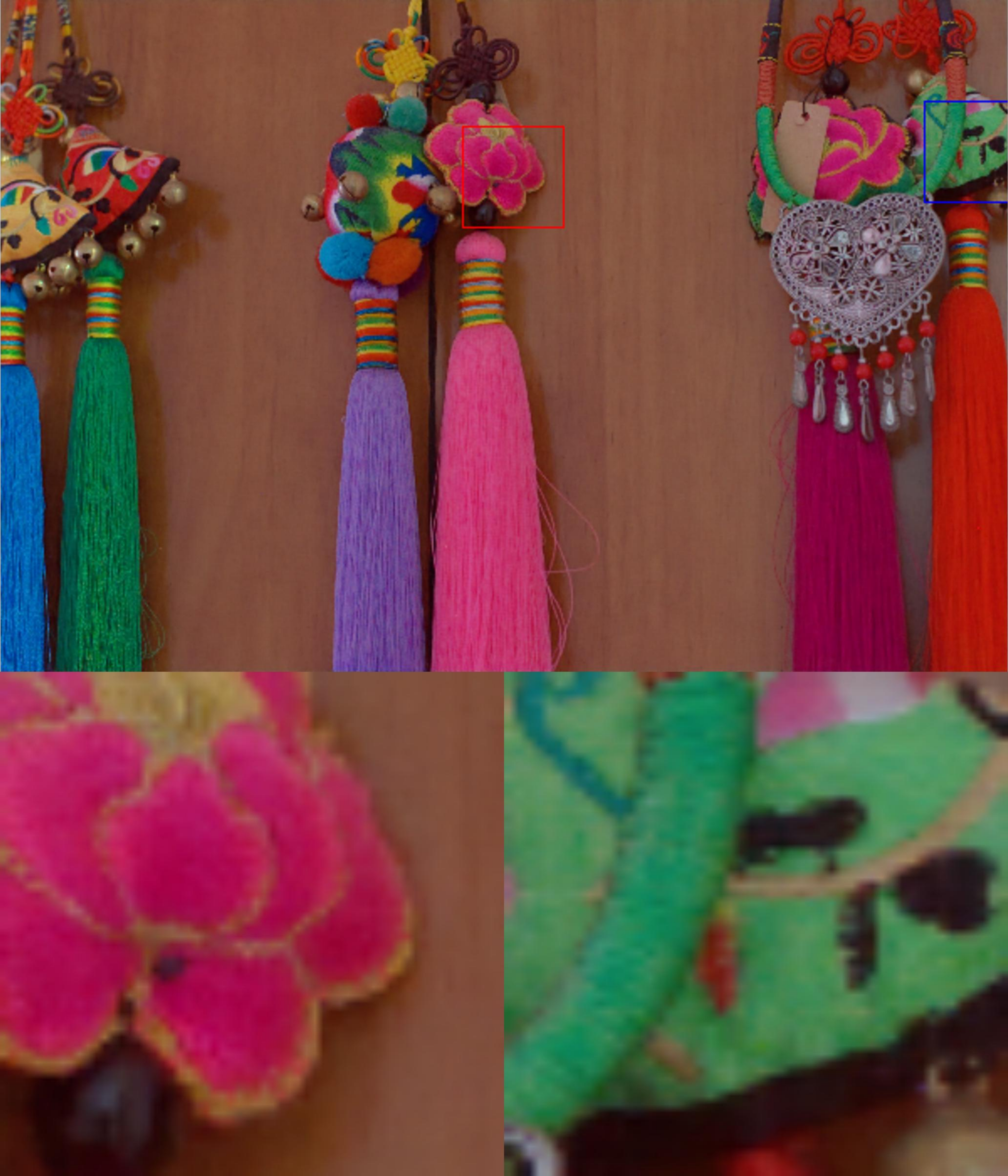}}
    \centering{\includegraphics[width=\textwidth]{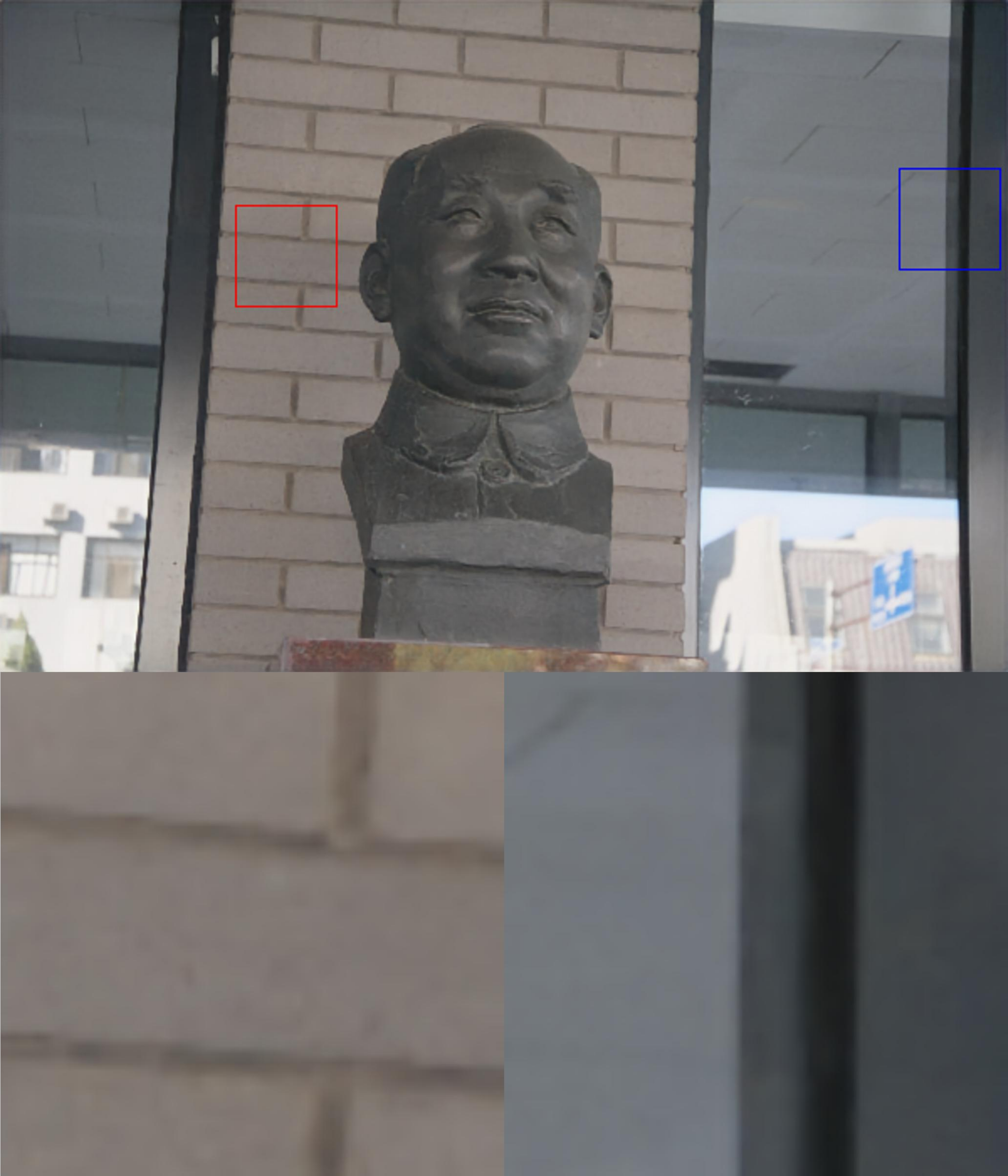}}
    \centering{\includegraphics[width=\textwidth]{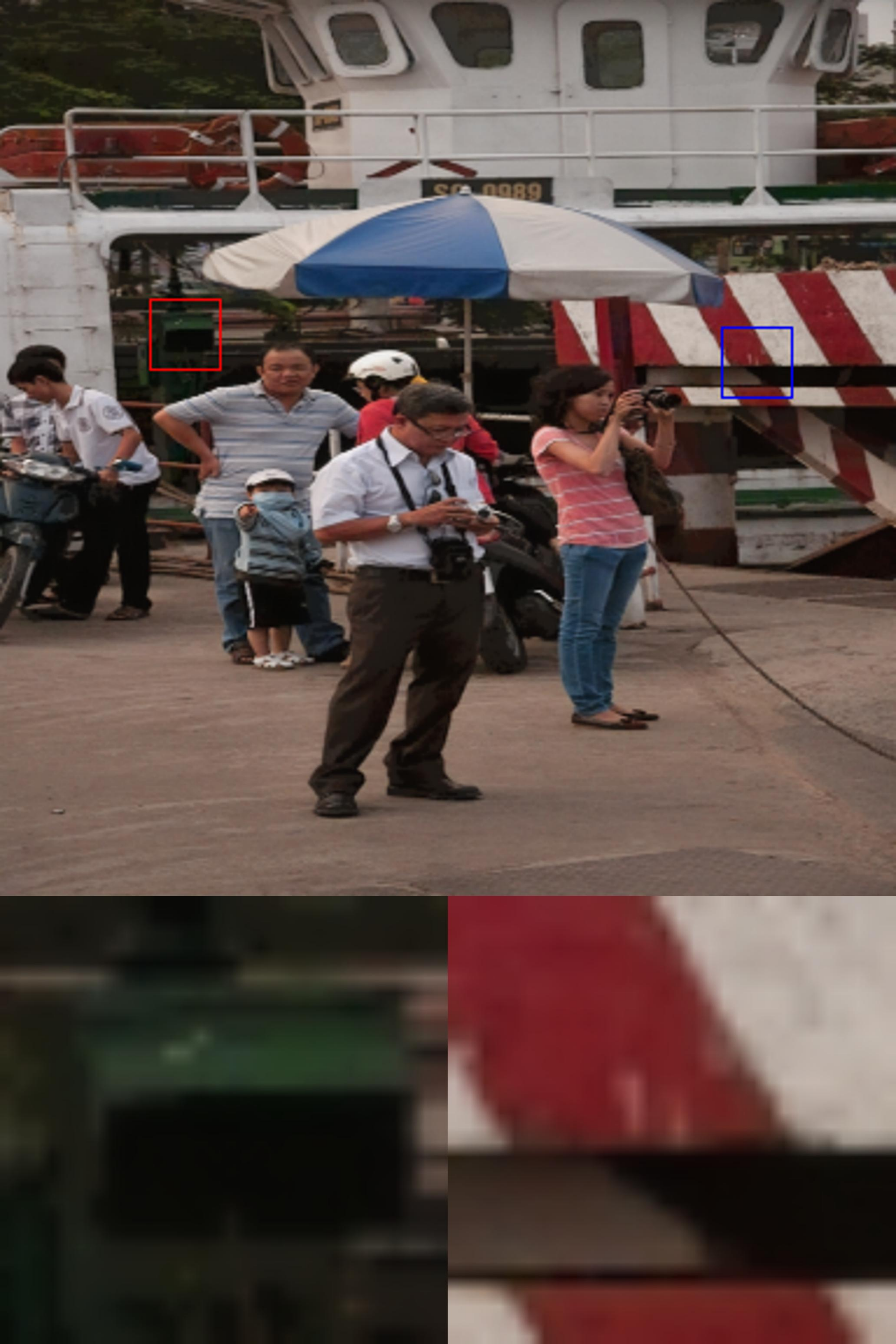}}
    \centerline{\small (g) Ours}
\end{minipage}
\begin{minipage}[t]{0.12\linewidth}
    \centering
    \vspace{3pt}
    \centering{\includegraphics[width=\textwidth]{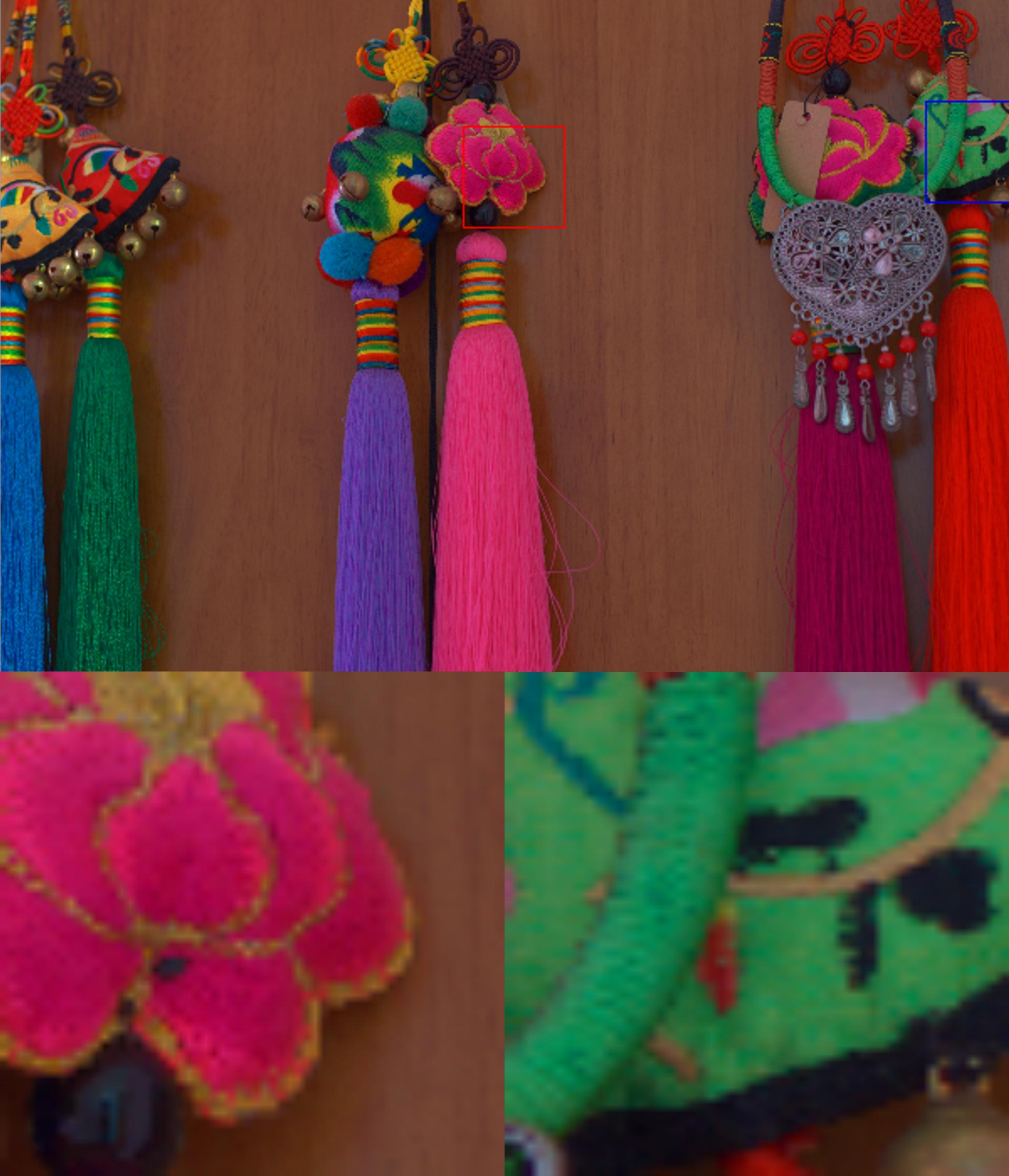}}
    \centering{\includegraphics[width=\textwidth]{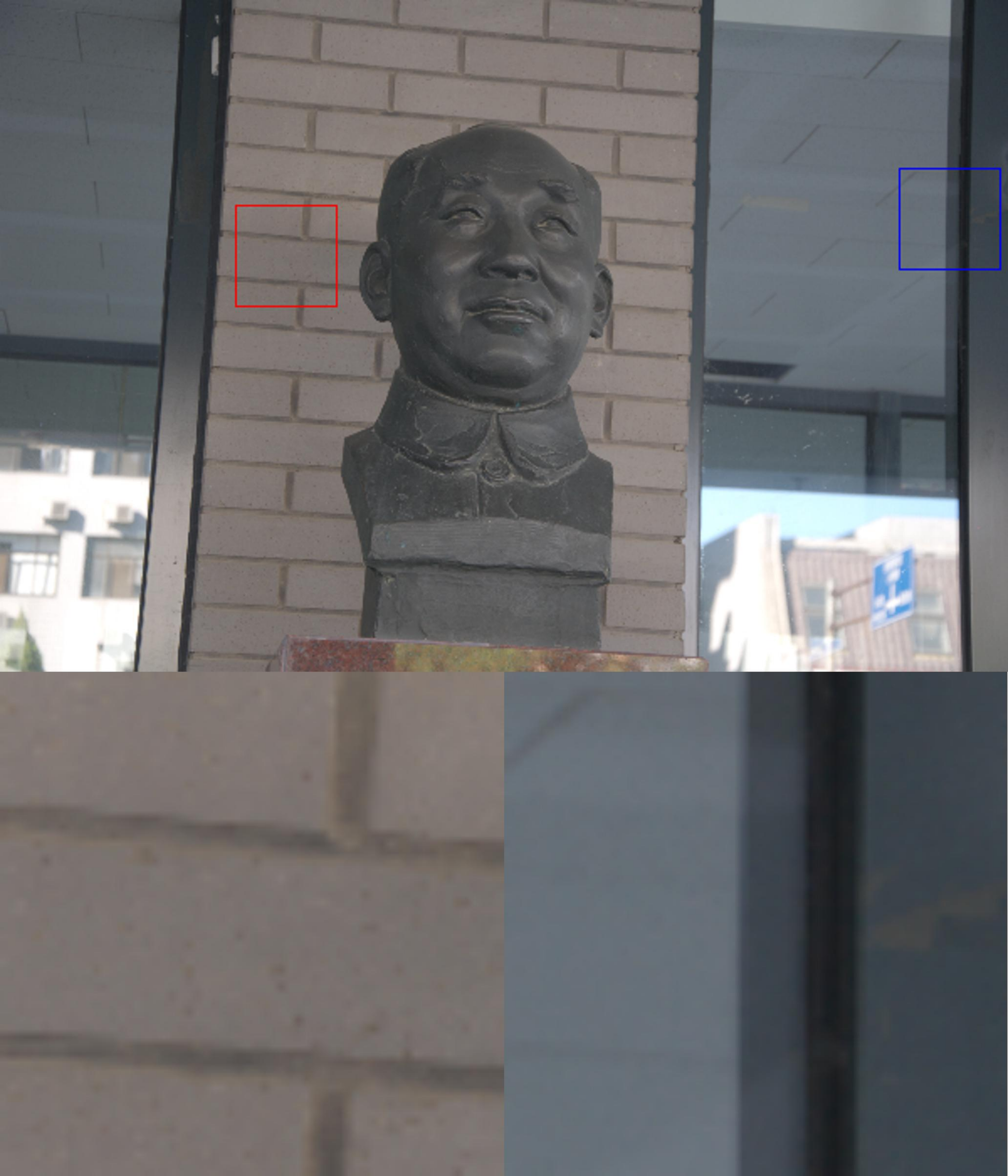}}
    \centering{\includegraphics[width=\textwidth]{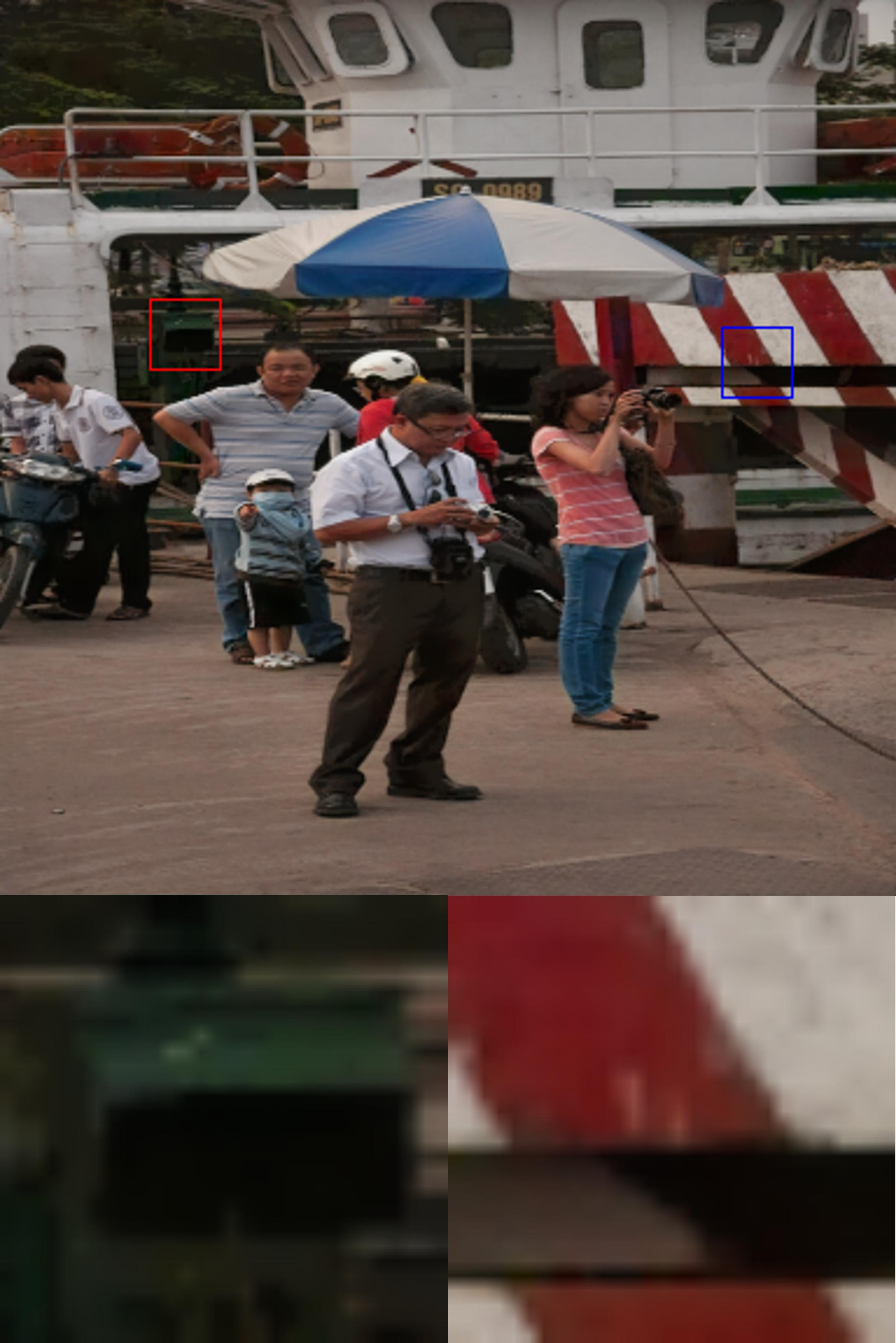}}
    \centerline{\small (h) GT}
\end{minipage}
\vspace{-0.2cm}
 \caption{Visual comparison of the enhanced images yielded by five different SOTA methods \cite{snr_net,cai2023retinexformer,hou2024global,yan2025hvi,bai2024retinexmamba} and proposed FusionNet on LOLv1 (top row), LOLv2-real (middle row), and LOLv2-synthetic (bottom row).}
 \label{fig:LOL}
\end{figure*}

\subsection{Qualitative Evaluations}
Quantitative metrics can indicate how closely a model's output approaches the ground truth (GT), but for low-light enhancement tasks, the subjective visual perception of the image is equally important. Therefore, we conducted qualitative experiments to compare our FusionNet with five other state-of-the-art (SOTA) methods. We selected a representative image from each of the three subsets of the LOL dataset and used small boxes to zoom in on two high-frequency detail regions in each image for comparison, as shown in Fig. \ref{fig:LOL}.
The results demonstrate that FusionNet not only closely matches the GT in terms of overall color, including vividness, contrast, brightness, and saturation, but also achieves excellent visual quality. FusionNet preserves high-frequency details, ensuring that structural information remains clear and visible.

Furthermore, we compared FusionNet with the two methods being fused, Retinexformer and CIDNet. As in Figs. \ref{fig:LOL} (c)and (f), compared to Retinexformer, the results from CIDNet are more vibrant, with higher contrast and clearer structures. However, CIDNet struggles with brightness stability: in LOLv1, it tends to be too bright, while in LOLv2-syn, it is noticeably too dark.
After fusion, we observed that FusionNet (see Fig. \ref{fig:LOL}(g)) effectively takes an average between these two methods, maintaining the strengths of both. It ensures that the overall output image's brightness and contrast fall within a comfortable range. This demonstrates the strong performance of our Linear Fusion, which can combine the advantages of multiple models to achieve superior image enhancement results.

\begin{table}
    \centering
    \small
    \caption{Ablations of four fusion strategies in LOLv1 dataset. The "S" denotes single-stage methods, which are fused in our FusionNet, and "F" represents different fusion methods. The best results are in \textcolor{red}{red}.}
    \resizebox{\linewidth}{!}{
    \begin{tabular}{l|l|ccc}
        \toprule
        &Methods &  PSNR$\uparrow$ & SSIM$\uparrow$ & LPIPS$\downarrow$ \\
        \midrule
        \multirow{3}{*}{S}& CIDNet &23.50 &\textcolor{red}{0.870} &0.105\\
        & RetinexFormer &25.15 &0.845 &0.131\\
        & ESDNet &22.44 &0.844 &0.109\\
        \midrule
        \multirow{4}{*}{F}&Serially Connect & 23.25 & 0.846 & 0.107   \\
        &Multi-stage Serial & 23.50 & 0.847 & 0.108 \\
        &Parallel-stage Serial &24.57 &0.853 & 0.124   \\
        &Linear Fusion (Ours)  & \textcolor{red}{25.17} & 0.857 & \textcolor{red}{0.103}  \\
        \bottomrule
    \end{tabular}
    }
    \label{tab:ablation_method}
    \vspace{-0.3cm}
\end{table}

\begin{figure*}[h]
\centering
\begin{minipage}[t]{0.11\linewidth}
    \centering
    \vspace{3pt}
    \includegraphics[width=\textwidth]{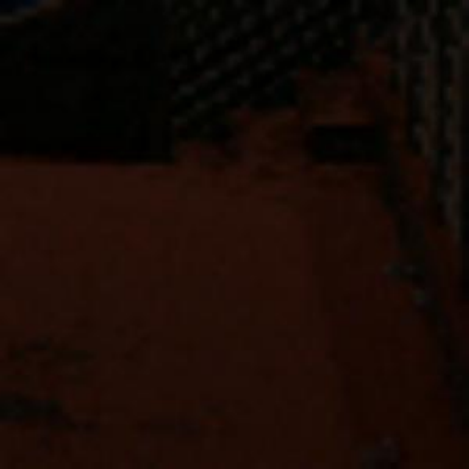}
    \includegraphics[width=\textwidth]{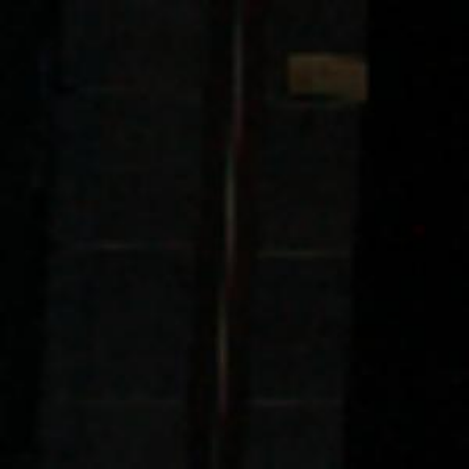}
    \centerline{\small Input}
\end{minipage}
\begin{minipage}[t]{0.11\linewidth}
    \centering
    \vspace{3pt}
    \includegraphics[width=\textwidth]{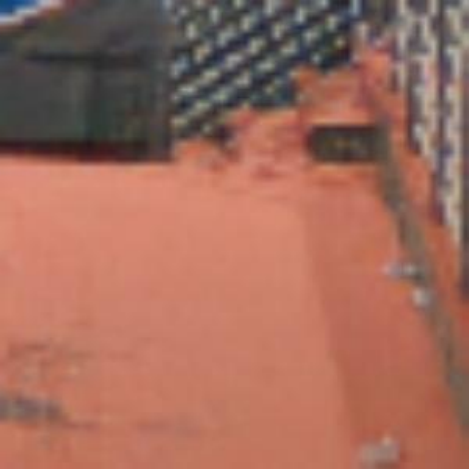}
    \includegraphics[width=\textwidth]{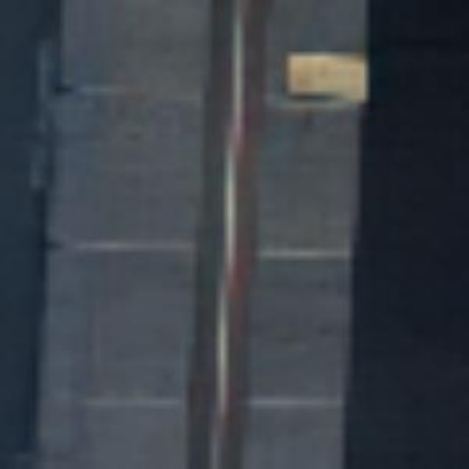}
    \centerline{\small Serially Connect}
\end{minipage}
\begin{minipage}[t]{0.11\linewidth}
    \centering
    \vspace{3pt}
    \includegraphics[width=\textwidth]{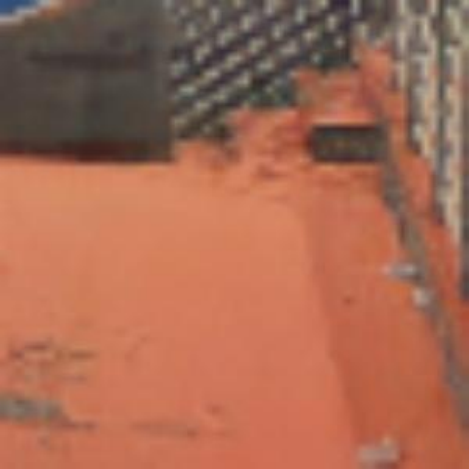}
      \includegraphics[width=\textwidth]{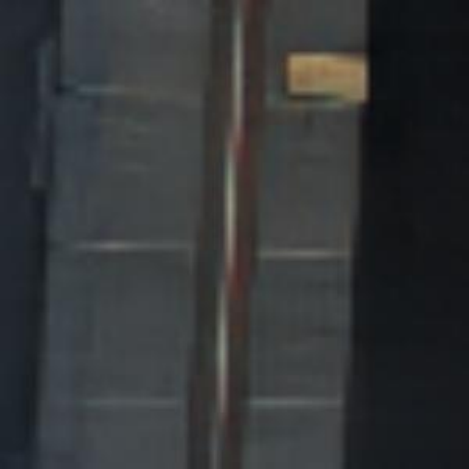}  
    \centerline{\small Multi-stage}
\end{minipage}
\begin{minipage}[t]{0.11\linewidth}
    \centering
    \vspace{3pt}
    \includegraphics[width=\textwidth]{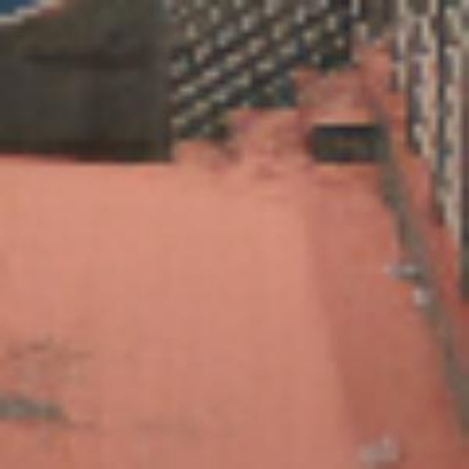}
    \includegraphics[width=\textwidth]{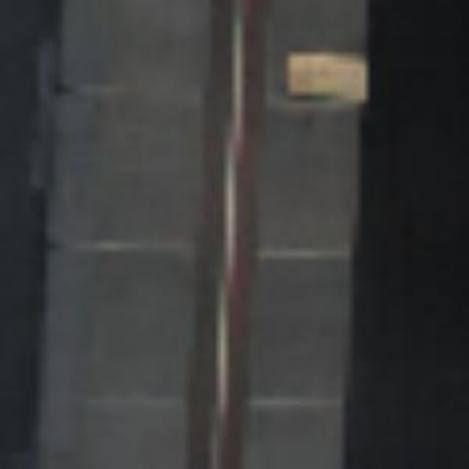}
    \centerline{\small Parallel-stage}
\end{minipage}
\begin{minipage}[t]{0.11\linewidth}
    \centering
    \vspace{3pt}
    \includegraphics[width=\textwidth]{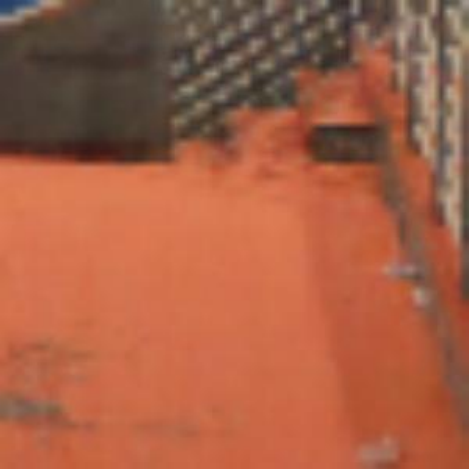}
    \includegraphics[width=\textwidth]{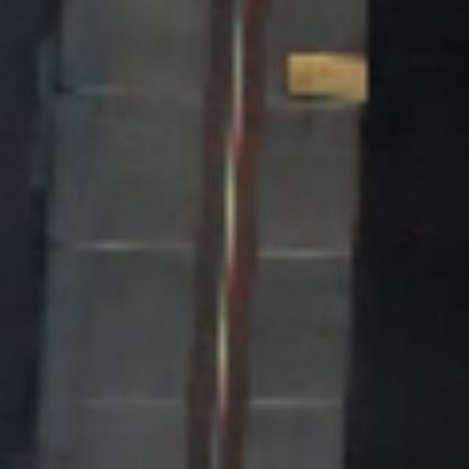}
    \centerline{\small Ours}

\end{minipage}
\begin{minipage}[t]{0.11\linewidth}
    \centering
    \vspace{3pt}
    \includegraphics[width=\textwidth]{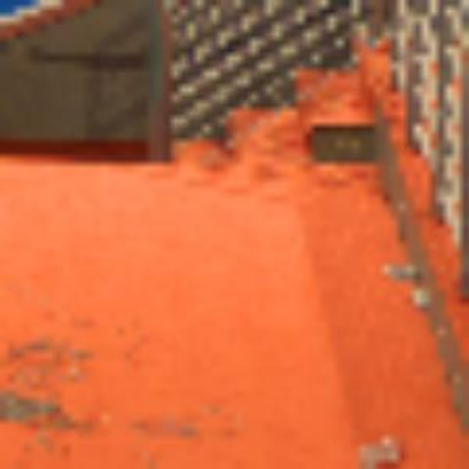}
    \includegraphics[width=\textwidth]{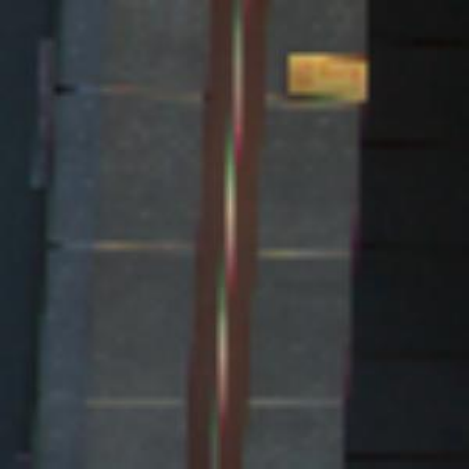}
    \centerline{\small GT}
\end{minipage}
\caption{Visualization of different fusion strategies of RetinexFormer \cite{cai2023retinexformer}, CIDNet \cite{yan2025hvi}, and ESDNet \cite{yu2022towards}.}
\label{fig:ab2}
\end{figure*}

\subsection{Ablation Study}
\textbf{Different Fusion Strategies.}
We conducted an ablation study focusing on the fusion strategy and individual modules within FusionNet on the LOLv1 dataset for fast convergence. First, analyzing the three methods in Section S of Tab. \ref{tab:ablation_method}, we observe that CIDNet achieves the highest SSIM and LPIPS scores but has a relatively low PSNR. In contrast, RetinexFormer compensates for this shortcoming by achieving higher PSNR; however, its LPIPS score is significantly lower, indicating a need for additional improvements in perceptual quality.
To address this limitation, we introduced ESDNet, aiming to enhance the overall perceptual quality. By leveraging the strengths of these three models, our goal is to fuse them into a more robust method that achieves superior overall performance.

Next, we integrated these three models into four different fusion strategies. In the Serially Connect network (see Section F of Tab. \ref{tab:ablation_method}), we sequentially combined RetinexFormer, CIDNet, and ESDNet into a single large network and retrained the entire model in a single stage. However, we found that this approach produced worse results than training the three networks separately.
The primary reason is that the networks interfere with each other, leading to conflicts that hinder effective learning. Additionally, this approach is highly susceptible to the small-parameter butterfly effect, where minor changes in the model can cause instability and even network collapse, making it difficult to converge to an optimal solution.

In the Multi-stage Serial network, we trained the models sequentially in the order of CIDNet, RetinexFormer, and ESDNet, freezing the parameters of the previous stage before proceeding to the next. To ensure a fair comparison, we kept the number of training iterations consistent across all strategies, allowing us to evaluate their performance under the same time constraints.
As shown in Tab. \ref{tab:ablation_method}, Multi-stage Serial improved PSNR by 0.25 dB compared to the Serially Connect structure, while SSIM and LPIPS remained largely unchanged. This result suggests that although this structure has the potential for better convergence, its performance is constrained by the limited number of iterations. Since the network requires significantly more iterations to reach its optimal state, its training cost is high, making it time-intensive and less efficient.

For the Parallel-stage Serial strategy, we first trained and tested RetinexFormer and CIDNet separately on LOLv1, obtaining their respective output images. We then concatenated these outputs along the channel dimension, forming a six-channel input image. Finally, we retrained ESDNet, using the six-channel image as input to generate the final output.
Experimental results (see Section F of Tab. \ref{tab:ablation_method}) show that this method outperforms the previous two in both PSNR and SSIM. However, LPIPS decreases, likely due to the lossy nature of ESDNet's feature extraction. 
Since ESDNet may struggle to extract optimal features from the concatenated inputs, it improves PSNR and SSIM, both of which rely on pixel-wise comparisons, but compromises perceptual quality, leading to a drop in LPIPS.

Finally, our Linear Fusion method achieves the best performance across all three metrics, as Tab. \ref{tab:ablation_method}. It not only surpasses the original single-stage models but also outperforms all other fusion strategies. This demonstrates that our fusion approach not only enhances the performance of individual methods but also provides the simplest and most effective solution among multi-model fusion strategies—achieving optimal results without requiring additional training.
The visual comparison in Fig. \ref{fig:ab2} further validates the superiority of Linear Fusion, showing that it achieves the most balanced brightness, vividness, color balance, and contrast. These results strongly support the effectiveness of our approach in producing high-quality enhanced images.

\begin{figure}
    \centering
    \includegraphics[width=1\linewidth]{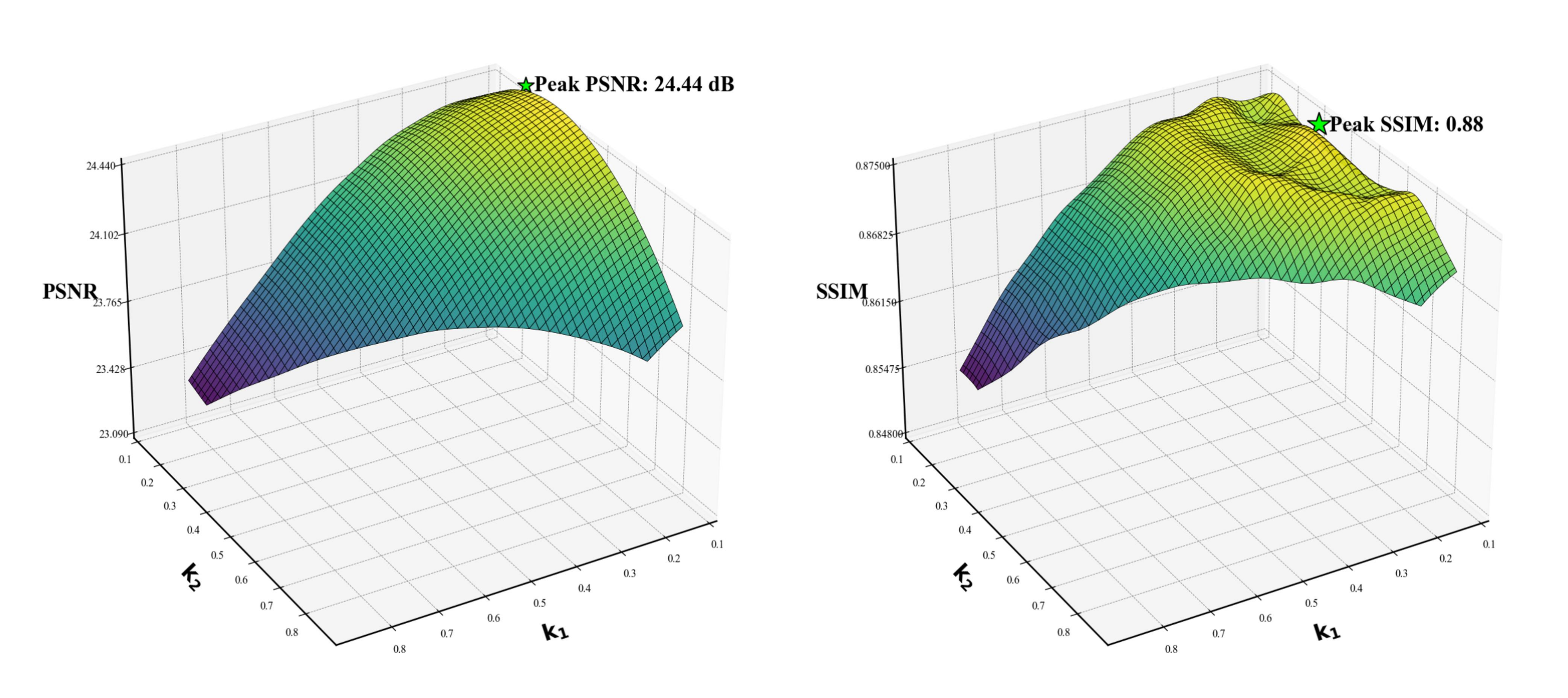}
    \caption{Three-dimention visualization of different $k_i$s on PSNR and SSIM in LOLv2-real dataset.}
    \label{fig:ab3}
\end{figure}

\textbf{Different parameter $k_i$s.} In our FusionNet, the linear fusion function has three customized parameters that followed $k_1+k_2+k_3=1$ and Eq. \ref{eq:1}.
As Fig. \ref{fig:ab3}, both PSNR and SSIM reached their peak at specific values of $k_1 = 0.16,k_2 = 0.40,k_3=0.44$, which counterparts RetinexFormer, CIDNet, and ESDNet, respectively. Interestingly, while PSNR exhibits a single global maximum, SSIM forms a group of local maxima. This intriguing phenomenon warrants further investigation, and we will explore its underlying causes in future research.

\section{Conclusion}
In this paper, we proposed FusionNet, a parallel framework based on multi-model linear fusion that effectively integrates the local feature extraction capabilities of CNNs, the global context modeling advantages of Transformers, and the luminance-color decoupling properties of the HVI color space. 
By introducing a linear fusion strategy guaranteed by Hilbert space theory, FusionNet avoids network collapse, significantly reduces training costs, and enhances model generalization. 
Experiments demonstrate that this method achieves state-of-the-art performance in the CVPR 2025 NTIRE challenge and across LOL benchmark datasets. 
Both quantitative metrics (\eg PSNR, SSIM) and qualitative visual results outperform existing methods. 
In future work, we will explore dynamic weight adjustment mechanisms to optimize fusion strategies for different scenarios and improve deployment efficiency in real-time applications.

\newpage
{
    \small
    \bibliographystyle{ieeenat_fullname}
    \bibliography{main}
}


\end{document}